\documentclass[final]{siamart171218}

\usepackage{color,soul}

\usepackage{algorithm,algpseudocode}

\usepackage{graphicx}

\usepackage{threeparttable}
\usepackage{enumitem}
\usepackage{multirow}
\usepackage{url}
\usepackage{bm}

\usepackage{changes}
\usepackage[all]{xy}
\usepackage{hyperref}
\usepackage{amssymb}
\usepackage{amsmath} 
\usepackage{multirow}
\usepackage{wrapfig}
\usepackage{caption}
\usepackage{subcaption}
\usepackage{bbm}

\newcommand{\eq}[1]{\begin{equation}\label{#1}}
  \newcommand{\en}{\end{equation}}
\def\up#1{^{(#1)}}%

\def\diag{\mbox{diag}}
\def\argmax{\mbox{argmax}}

\def\inv{^{-1}}
\def\RR{\mathbb{R}}

\def\calS{\mathcal{S}} 
\def\nref#1{(\ref{#1})}

\usepackage{bbm}

\newcommand{\ones}{\mathbf{1}}
\graphicspath{{./FIGS/}}

\DeclareMathOperator{\softmax}{softmax}
\DeclareMathOperator{\relu}{ReLU}

\title{Graph coarsening: From scientific computing to
  machine learning} 

\date{\today}

\author{Jie Chen\thanks{MIT-IBM Watson AI Lab, IBM Research.
    E-mail: chenjie@us.ibm.com. Work supported by DOE award DE-OE0000910.}
  \and Yousef Saad\thanks{University of Minnesota.
    E-mail: \{saad, zhan5260\}@umn.edu. Work supported by NSF grant
    DMS-2011324. The authors are listed alphabetically.} 
  \and Zechen Zhang\footnotemark[2]
}

\begin{document}

%%\tableofcontents

\maketitle

\begin{abstract}
The general method of graph coarsening or graph reduction has been a
remarkably useful and ubiquitous tool in scientific computing and it
is now just starting to have a similar impact in machine learning.
The goal of this paper is to take a broad look into coarsening
techniques that have been successfully deployed in scientific
computing and see how similar principles are finding their way in more
recent applications related to machine learning.  In scientific
computing, coarsening plays a central role in algebraic multigrid
methods as well as the related class of multilevel incomplete LU
factorizations.  In machine learning, graph coarsening goes under
various names, e.g., graph downsampling or graph reduction. Its goal
in most cases is to replace some original graph by one which has fewer
nodes, but whose structure and characteristics are similar to those of the
original graph. As will be seen, a common strategy in these methods is
to rely on spectral properties to define the coarse graph.
\end{abstract}



\section{Introduction}\label{sec:introduction}
The idea of `coarsening,' i.e., exploiting a smaller set in place of a
larger or `finer' set has had numerous uses across many disciplines of
science and engineering. The term `coarsening' employed here is
prevalent in scientific computing, where it refers to the usage of 
coarse meshes to solve a given problem by, e.g., multigrid (MG) or
algebraic multigrid (AMG) methods. On the other hand, the terms `graph
downsampling,' `graph reduction,' `hierarchical methods,' and
`pooling' are common in machine learning.  Similarly, the related idea
of clustering is an important tool in data-based applications.  Here,
the analogous term employed in scientific computing is `partitioning.'
These notions---graph partitioning, clustering, coarsening---are
strongly inter-related.  It is possible to use partitioning for the
task of clustering data, by first building a graph that models the
data which we then partition. Also, coarsening plays an important role
in developing effective graph partitioning methods. Further, note that
it is possible to partition a graph by just finding some clustering of
the nodes, using a method from data sciences such as the K-means
algorithm.

In scientific computing, the best known instance of coarsening
techniques is in MG and AMG methods
\cite{Hackbusch-1985,Ruge-Stuben-AMG,MG-book,MG-tutorial}.  Classical
MG methods started with the independent works of Bakhvalov
\cite{Bakhvalov} and Brandt \cite{Brandt-77}. The important discovery
revealed by these pioneering articles is that relaxation methods for
solving linear systems tend to stall after a few steps, because they
have difficulty in reducing high-frequency components of the error.
Because the eigenvectors associated with a coarser mesh are direct
restrictions of those on the fine mesh, the idea is to project the
problem into an `equivalent' problem on the coarse mesh for error
correction and then interpolate the solution back into the fine level.
This basic 2-level scheme can be extended to a multilevel one in a
variety of ways. MG does not use graph coarsening specifically because
it relies on a mesh and it is more natural to define a coarse mesh
using processes obtained from the discretization of the physical domain.
On the other hand,
AMG aims at general problems that do not necessarily have a mesh
associated with them.  For AMG, the graph representation of the
problem at a certain level is explicitly `coarsened' by using various
mechanisms~\cite{Ruge-Stuben-AMG,MG-tutorial,Saad-book2}.  Since these
mechanisms are geared toward a certain class of problems, essentially
originating from Poisson-like partial differential equations,
researchers later sought to extend AMG ideas in order to define
algebraic techniques based on incomplete LU (ILU) factorizations
\cite{Bank-Wagner-MLILU,HEM,IMF,AxPo:robust88,Axelsson-Larin,Axelsson-Neytcheva,
  Axelsson-Vassilevski-a,Axelsson-Vassilevski-b,Axelsson-Vassilevski-c,MRILU}.

One can say that the idea of coarsening a graph in data-related
applications started with the 1939 article of Kron~\cite{kron39},
whose aim was to downsample electrical networks.  Kron used his deep
intuition to define coarsening techniques that rely on Schur
complements, with the goal of obtaining sparse graphs.  The
justifications for the proposed technique were based on intuition
rooted on knowledge about electrical networks. The related technique,
widely known as Kron reduction, was revived by D\"orfler and Bullo
\cite{Kron-paper13} who provided a more rigorous theoretical
foundation.  Later Shuman et al. \cite{MSkron} extended the Kron reduction
into a multilevel framework. In parallel with this line of work, a
number of authors developed techniques that bypassed the need to form
or approximate the Schur complement relying instead on node
aggregation and matching
\cite{hermsdorff2019unifying,jin2020graph,MILE-paper18,HARP-paper18,LINE,HRFangEtAlcikm1,HRFangEtAlcikm2,hss:mlevel-08}.

Applications of graph coarsening in machine learning generally fall in
two categories. First, coarsening is instrumental in graph
embeddings. When dealing with learning tasks on graphs, it is very
convenient to represent a node with a vector in $\RR^d$ where $d$ is
small. The mapping from a node to the representing vector is termed
\emph{node (or vertex) embedding} and finding such embeddings tends to be
costly. Hence, the idea is to coarsen the graph first, perform some
embedding at the coarse level, and then refine-propagate the embedding
back to the upper level; see
\cite{HARP-paper18,MILE-paper18,GraphZoom20,PanayotGraphEmbed21} for
examples of such techniques.  The second category of applications is
when invoking \emph{pooling} on graphs, in the context of graph neural
networks (GNNs)
\cite{DiffPool-paper18,SortPool_paper18,Ma2021}. However, in the
latest development of GNNs, coarsening is not performed on the given
graph at the outset. Instead, coarsening is part of the neural network
and it is \emph{learned} from the  data. Another class of applications of
coarsening is that of graph filtering, as illustrated by the articles
\cite{MSkron,shuman-survey}.

The goal of this paper is to show how the idea of coarsening has been
exploited in scientific computing and how it is now emerging in
machine learning.  While the problems under consideration in
scientific computing are fundamentally different from those of machine
learning, the basic ingredients used in both methods are striking by
their similarity.  The paper starts with a discussion of graph
coarsening in scientific computing (Section \ref{sec:SC}), followed by
a section on graph coarsening in machine learning (Section
\ref{sec:ML}). We also present some newly developed coarsening methods
and results, in the context of machine learning, in Sections
\ref{sec:spectral}--\ref{sec:LESC.s}.

%%%%%%%%%%%%%%%%%%%%%%%%%%%%%%%%%%%%%%%%%%%%%%%%%%%%%%%%%%%%
\subsection{Notation and preliminaries}\label{sec:Notation}
We denote by $G = (V,E)$ a graph with $n$ nodes and $m$ edges, where
$V$ is the node set and $E$ is the edge set. The weights of the edges
of $G$ are stored in a matrix $A$, so $a_{ij}$ is the weight of the
edge $(i,j) \in E$.  In most cases we will assume that the graph is
undirected. We sometimes use $G=(V,E,A)$ to denote the graph, when $A$
is emphasized.

The sum of row $i$ of $A$ is called the degree of node $i$ and the
diagonal matrix of the degrees is called the degree matrix:
\eq{eq:defs}
d_i = \sum_{j=1}^n a_{ij};\quad
D = \diag ( d_i ) . 
\en
With this notation, the graph Laplacian matrix $L$ is defined as:
\begin{equation} \label{eq:Glap} L = D-A .  \end{equation}
This definition implies that $L \ones = 0 $ where $\ones$ is the vector of all ones;
i.e., $\ones$ is an eigenvector associated with the  eigenvalue zero.
In the simplest case, each weight $a_{ij}$ is either zero (not
adjacent) or one (adjacent).  A simple, yet very useful, property of
graph Laplacians is that for any vector $x$, we have the relation
\eq{eq:LaplProp}
x^T L x = \sum_{ij} a_{ij} |x_i - x_j|^2 \ . 
\en

The normalized Laplacian is defined as follows:
\eq{eq:NormLap}
\widehat{L}  = D^{-1/2} L D^{-1/2} = I - D^{-1/2} A D^{-1/2} \ .
\en
Note that the diagonal entries of $\widehat{L}$ are all ones.  The
matrix is again singular and has the null vector $D^{1/2} \ones$.

We will also make use of the incidence matrix denoted by
$B\in\mathbb{R}^{n\times m}$. A column $b_e$ of $B$ represents an edge
$e\in E$ between nodes $i$ and $j$ with weight $a_{ij}$, and its
$k$-th entry is defined as follows:
\begin{equation}\label{eq:b}
    b_e(k) = 
\begin{cases}
    +\sqrt{a_{ij}}, & k=i,\\
    -\sqrt{a_{ij}}, & k=j,\\
    0,  & \text{otherwise}.
\end{cases}
\end{equation}
Note that the two nonzero values of $b_e(k)$ have opposite signs, but
we have a choice regarding which of $i$ and $j$ is assigned the
negative sign. Unless otherwise specified, we simply assign the
negative sign to the smaller of $i$ and $j$. As is well-known, the
graph Laplacian can be defined from the incidence matrix through the
relation $L = BB^{T}$.

%%%%%%%%%%%%%%%%%%%%%%%%%%%%%%%%%%%%%%%%%%%%%%%%%%%%%%%%%%%%
\subsection{Terminology and notation specific to machine learning}\label{sec:notation.ml}
In data-related applications, graph nodes are often equipped with
feature vectors and labels. We use an $n\times d$ matrix $X$ to denote
the feature matrix, whose $i$-th row is the feature vector of node
$i$. We use an $n\times c$ matrix $Y$ to denote the label matrix,
where $c$ is the number of categories. When a node $i$ belongs to
category $j$, $Y_{ij}=1$ while $Y_{ij'}=0$ for all $j'\ne j$. Each row
of $Y$ is called a `one-hot' vector. When there are only two categories,
the $n\times2$ matrix $Y$ can equivalently be represented by an
$n\times1$ binary vector $y$ in a straightforward manner.

For example, in a transaction graph, where nodes represent account holders
and edges denote transactions between accounts, a node may have $d=4$
features: account balance, account active days, number of incoming
transactions, and number of outgoing transactions; as well as $c=3$
categories: individual, non-financial institution, and financial
institution. A typical task is to predict the account category given the features.

The feature matrix $X$ provides complementary information to a graph
$G=(V,E,A)$ that captures relations between data items.
One should not confuse the feature matrix with a data
matrix, which by convention clashes with the notation $X$. Let us use
$Z$ instead to denote a data matrix, whose $i$-th row is denoted by
$z_i$. One may construct the graph $G$ from $Z$. For example, in a
$k$-nearest neighbors ($k$NN) graph, there is an edge from node $i$ to
node $j$ if and only if $j$ is an index of the element among the $k$
smallest elements of $\{r_{ij}=\|z_j-z_i\|\mid j\ne i\}$. One may even
define the weighted adjacency matrix $A$ as $a_{ij}=e^{-r_{ij}}$ when
there is an edge $(i,j)$ and $a_{ij}=0$ otherwise. In this case, the
constructed graph is entirely decided by the data matrix $Z$, rather
than by holding complementary information to it, as is done with  the feature matrix.

\section{Graph coarsening in scientific computing}\label{sec:SC}
Given a graph $G=(V,E)$, the goal of graph coarsening is to find a
smaller graph $G_c=(V_c,E_c)$ with $n_c$ nodes and $m_c$ edges, where 
 $n_c < n$, which is a good approximation of $G$ in some
sense. Specifically, we would like the coarse graph to provide a
faithful representation of the structure of the original graph.  We
denote the adjacency matrix of $G_c$ by $A_c$ and the graph Laplacian
of $G_c$ by $L_c$.

\begin{figure}[h]
\centerline{\includegraphics[width=0.6\textwidth]{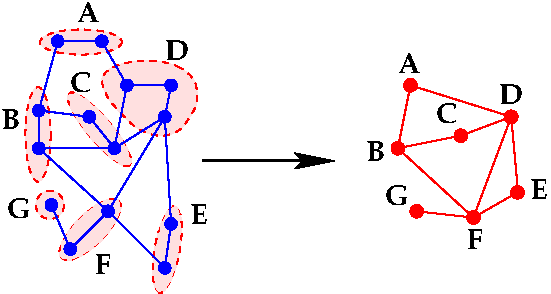}}
\caption{Coarsening a graph.}
\end{figure} 

We will first elaborate on one of the most important scenarios that invoke
coarsening (Section~\ref{sec:AMG}) and then discuss several
representative approaches to it (Sections~\ref{sec:HEM}
to~\ref{sec:coarse.related}). Note that in practice, coarsening often
proceeds recursively on the resulting graphs; by doing so, we obtain a
hierarchy of approximations to the original graph.

%%%%%%%%%%%%%%%%%%%%%%%%%%%%%%%%%%%%%%%%%%%%%%%%%%%%%%%%%%%%
\subsection{Multilevel methods for linear systems: AMG and multilevel ILU}
\label{sec:AMG}
Graph coarsening strategies are usually invoked when solving linear systems of
equations, by multilevel methods such as (A)MG
\cite{Hackbusch-1985,MG-book,Ruge-Stuben-AMG} or Schur-based multilevel
techniques \cite{Chen-book, Saad-Suchomel-ARMS, HEM,McLSaa:06, McLSaa06NONS:07,
  Bank-Wagner-MLILU, Axelsson-Vassilevski-a}.  In (A)MG, this amounts to
selecting a subset of the original (fine) grid, known as the `coarse grid.'  In
AMG, the selection of coarse nodes is made in a number of different ways.  The
classical Ruge-St\"uben strategy \cite{Ruge-Stuben-AMG} selects coarse nodes
based on the number of `strong connections' that a node has. Here, nodes $i$ and
$j$ are strongly connected if $a_{ij}$ has a large magnitude relative to other nonzero
off-diagonal elements or row $i$.  The net effect of this strategy is that each
fine node is strongly coupled with the coarse set.  In other methods, the
strength of connection is defined from the speed with which components of a
relaxation scheme for solving the homogeneous system $Au = 0$ converge to zero;
see \cite{brandt-review-01,Safro-2011,ChenSafro-2011} and Section \ref{sec:ALG}
for additional details.  \iffalse Other techniques (compatible relaxation, ...)
rely on the speed with which components of a relaxation scheme for solving the
homogeneous system $Au = 0$ converge to zero.  For (A)MG, the coarse points are
selected so that the residual equation $Ae = r$, for some error $e$ and residual
$r$, is slow to converge; see, e.g., \cite{MG-tutorial}.  \fi For multilevel
Schur-based methods, such as multilevel ILU, the coarsening strategy may
correspond to selecting from the adjacency graph of the original matrix, a
subset of nodes that form an independent set \cite{Saad-Suchomel-ARMS}, or a
subset of nodes that satisfy good diagonal dominance properties \cite{saad-ddPQ}
or that limit the growth in the inverse LU factors of the ILU factorization
\cite{Bollhoefer-robilu, Bollhoefer-AMILU}.

The coarsening strategy can be expanded into a multilevel framework by
repeating the process described above on the graph associated with the
nodes in the coarse set.  Let $G_0$ be the original graph $G$ and let
$G_1, G_2, \ldots, G_m$ be a sequence of coarse graphs such that
$G_\ell = (V_\ell, E_\ell)$ is obtained by coarsening on $G_{\ell-1}$
for $1\leq \ell < m$.  Let $A\up{0} \equiv A$ and $A\up{\ell}$ be the
matrix associated with the $\ell$-th level.  The graph $G_{\ell}$
admits a splitting into coarse nodes, $C_{\ell}$, and fine nodes,
$F_{\ell}$, so that the linear system at the $\ell$-th level, which
consists of the matrix $A\up{\ell}$ and the right-hand side
$f\up{\ell}$ can be reordered as follows:
\begin{equation}\label{eq:2}
A\up{\ell} = 
  \left[\begin{array}{cc}
A\up{\ell}_{CC} & A\up{\ell}_{CF} \\
        A\up{\ell}_{FC} & A\up{\ell}_{FF}\end{array}\right] \ , \quad
    f\up{\ell} =
    \begin{bmatrix}       f\up{\ell}_{C} \\ f\up{\ell}_{F} \end{bmatrix} . 
\end{equation}
Note that it is also possible to list the fine nodes first followed by
the coarse nodes; see \cite{HEM}.  The coarser-level graph
$G_{\ell+1}$ as well as the new system consisting of the matrix
$A^{(\ell+1)}$ and the right-hand side $f^{(\ell+1)}$ at the next
level, are constructed from $G_{\ell}$ and $A\up{\ell}$.  These are
built in a number of different ways depending on the method under
consideration.  For the graph, we can for example set two coarse nodes
to be adjacent in $G_{\ell+1}$, if their representative children are
adjacent in $G_{\ell}$. One common way to do this is to define two
coarse nodes to be adjacent in $G_{\ell+1}$ if they are parents of
adjacent nodes in $G_{\ell}$.  Next we discuss coarsening in the
specific context of AMG.

\subsubsection{Algebraic multigrid}
\begin{figure}
  \centerline{\includegraphics[width=.45\textwidth]{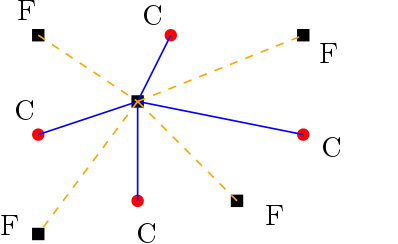}}
  \caption{A fine node and its nearest neighbors. Fine nodes are
    represented by a square and coarse ones by a disk. In the coarsening process
    the central, fine node is expressed as a combination of its coarse neighbors.}
  \label{fig:coars1}
\end{figure}

AMG techniques are all about generalizing the
\emph{interpolation} and \emph{restriction} operations of standard MG.
The coarsening process identifies for each fine node a set of nearest
neighbors from the coarse set. Using various arguments on the strength
of connection between nodes, AMG expresses a fine node $i$ as a linear
combination of a selected number of nearest neighbors that form a set
$C_i$, see Figure~\ref{fig:coars1}.  To simplify notation, we consider
only one level of coarsening and drop the subscript $\ell$.

If $C$ is the set of coarse nodes and $F$ is the set of fine nodes, we
can define related subspaces $\mathcal{X}_C$ and $\mathcal{X}_F$ of the
original space $\mathcal{X}=\RR^n$. In fact we can write $\mathcal{X}
= \mathcal{X}_C \oplus \mathcal{X}_F$.  Then, given a vector $x$ with
components in the coarse space $\mathcal{X}_C$, we associate a vector
$Px$ in the original space $\mathcal{X}$, whose $i$-th component is
defined as follows~\cite[13.6.2]{Saad-book3}:
\eq{eq:Pell}
[ P x ]_i  =
\left\{ \begin{array}{ll}
          x_i & \mbox{if}  \ i \ \in C, \\
          \sum_{j\in C_i}  p_{ij} x_j  & \mbox{otherwise} .
        \end{array} \right.
\en
      
The mapping $P$ sends a point of $\mathcal{X}_C$ into a point $y = Px$
of $\mathcal{X}$.  The value of $y$ at a coarse point, a node in $C$,
is the same as its starting value. The value at another node, one in
$F$, is defined from interpolated values at a few coarse points.
Thus, $P $ is known as the \emph{interpolation} operator.

The transpose of $P$ represents the \emph{restriction}, or coarsening
mapping.  In the context of AMG, it projects a point in $\mathcal{X}$
into a point in $\mathcal{X}_C$.  Each node in $C$ is a linear
combination of nodes of the original graph.

If we now return to the multilevel case where $P_{\ell}$ denotes a
corresponding interpolation operator at the $\ell$-th level, then AMG
defines the linear system at the next level using
\emph{Galerkin projection,} 
where the matrix and right-hand side are, respectively,
\eq{eq:AMGsys}
A^{(\ell+1)}= P_{\ell}^TA\up{\ell}P_{\ell}, \quad
f\up{\ell+1} = P_{\ell}^T f\up{\ell}  \ . 
\en
Recall that we started with the original system $A\up{0} x = f\up{0}$,
which corresponds to $\ell = 0$.  AMG methods rely on a wide variety
of iterative procedures that consist of exploiting different levels
for building an approximate solution.  It is important to note here
that \emph{the whole AMG scheme depends entirely on defining a
sequence of interpolation operators $P_\ell$ for $\ell = 0,
1,\ldots$}  Once the $P_{\ell}$'s are defined, one can design
various `cycles' in which the process goes back and forth from the
finest level to the coarsest one in an iterative procedure.

 When defining the interpolation operator $P_{\ell}$, there are two possible
extremes worth noting, even though these extremes are not used in AMG
in practice.  On the one end, we find the \emph{trivial interpolation} in
which the $p_{ij}$'s in equation
\nref{eq:Pell} are all set to zero. In this case, referring to \nref{eq:2},
$A\up{\ell+1}$ is simply $A\up{\ell+1} = A_{CC}\up{\ell}$.

The other extreme is the \emph{perfect interpolation}  case which yields the
Schur complement system. Here, the interpolation operator is
\[
  P_{\ell} = \begin{bmatrix}    
    - [A_{CC}\up{\ell}]\inv A_{CF}\up{\ell}
     \\ I 
  \end{bmatrix} .
\]
The right reduced matrix $A\up{\ell} P_{\ell} $ involves the Schur
complement matrix associated with the coarse block:
\[
  A\up{\ell} P_{\ell} = 
  \begin{bmatrix}
    O \\
    S_\ell
\end{bmatrix} \quad \mbox{where} \quad 
    S_\ell  = 
  A_{FF}\up{\ell} - A_{FC}\up{\ell}[ A_{CC}\up{\ell} ]\inv A_{CF}\up{\ell} . 
\] 
Then clearly,
\[
  A\up{\ell+1} = P_{\ell}^T A\up{\ell} P_{\ell} = S_\ell . 
\] 

Since this is the exact Schur complement, if we were to solve the
related linear system, we could recover the whole solution of the
original system by substitution. This approach is costly but there are
practical alternatives discussed in the
literature \cite{Bollhoefer-robilu,IMF,Bank-Wagner-MLILU,MayerCrout07}
that are based on Schur complements. However,
it is worth pointing out that, viewed from this angle, \emph{the goal
of all AMG methods is essentially to find inexpensive approximations
to the Schur complement system.}

\subsubsection{Multilevel ILU preconditioners based on coarsening}
The issue of finding a good ordering for ILU generated a great deal of
research interest  in the past; see, e.g.,
\cite{beszdu99,Benzi-Tuma-Order-00,Benzi-JM-99,bridson-tang99,clift-tang,
  doi-lichnewsky-91,DAzFoTa:unstructured92,Tang-rcm-anal,saad-ddPQ,Chow-Saad-stab}.
A class of techniques presented in \cite{HEM} consisted of
preprocessing the linear system with an ordering based on
coarsening. Thus, for a one-level ordering the matrix is ordered as
shown in \nref{eq:2}.  Then in a second level coarsening,
$A_{22}\up{0}$ is in turn reordered and we end up with a matrix
like:\footnote{For notational simplicity, for the subscripts of $A$ we
use 1 in place of $C$ and 2 in place of $F$.}
\begin{equation*}
  \left[\begin{array}{c|c}
         A_{11}\up{0} & A_{12}\up{0} \\ \hline
          A_{21}\up{0} &
            \begin{array}{c|c}
            A_{11}\up{1} & A_{12}\up{1} \\ \hline
            A_{21}\up{1} & A_{22}\up{1} \end{array} 
        \end{array}\right] .
%%       \label{eq:Mul2}.
\end{equation*}
This is repeated with $A_{22}\up{1}$ and further down for a few
levels.  Then the idea is simply to perform an ILU factorization of
the resulting reordered system.  Next, we describe a method based on
this general approach.
 
The first ingredient of the method is to define a weight $w_{ij} $ for
each nonzero pair $(i,j)$.  This will set an order in which to visit
the edges of the graph.  The strategies described next are `static' in
that given a certain matrix $A$ (one of the $A\up{\ell}$'s), these
weights are precomputed, in contrast with dynamic ones used in, e.g.,
\cite{Saad-Suchomel-ARMS,MacLach-al-2011,saad-ddPQ}.
If $n_i $ is the number of nonzero entries of row $i$ and $m_j$ is the number of
nonzero entries of column $j$, we define the
weights as follows:
\begin{eqnarray} 
w_{ij} &=& \min \left\{ \frac{|a_{ij}| }{\delta_r (i) } \ , \ \frac{|a_{ij}|
           }{\delta_c (j) } \right\} \ \mbox{where:} \label{eq:wgt}  \\
\delta_r (i) &= &\ \frac{\| A_{i,:} \|_1 }{n_i}
                  \quad \mbox{and} \quad \delta_c (j) \ = \ \frac{\| A_{:,j} \|_1}{m_j}
                  \label{eq:wgt1}
\end{eqnarray} 
where matlab notation is exploited and $\| \cdot \|_1$ is the usual
1-norm.  The two terms in the brackets of \nref{eq:wgt} represent the
importance of $|a_{ij}|$ relative to the other elements in the same
row and column, respectively.  If our goal is to put large entries in
the (1,2) block of the matrix when it is permuted (block
$A_{CF}\up{\ell}$ in~\nref{eq:2}), then we need to traverse
the graph starting from the largest to smallest $w_{ij}$.

The above defines an order in which to visit edges.  Next, each time
an edge $(i,j)$ is visited we need to determine which one of $i$ and
$j$ will be selected as a coarse node. This requires a `preference'
measure, or weight, for each node.  When $a_{kk} \ne 0$ we define the
impact of `pivot' $k$ as the average potential fill-in created when
eliminating unknown $k$. In the formula $a_{ij} = a_{ij} -
a_{ik}\times a_{kj}/a_{kk}$ employed in the $k$-th step of Gaussian
elimination, the term $-a_{ik}\times a_{kj} / a_{kk} $ is a potential
fill-in.  This is a very crude approximation because it assumes that
the entries have not changed.  We define the `impact' of the diagonal
entry $k$ as the inverse of the quantity:
\eq{eq:sigk}
 \sigma_k  =  \frac{ | a_{kk} | }{\delta_r (k) \delta_c (k) }  \ . 
\en
When visiting edge $(k,l)$, we add $k$ to the coarse set if
$\sigma_k>\sigma_l$ and $l$ otherwise.

\begin{figure}[h] 
  \begin{center}
  \includegraphics[height=0.41\textwidth]{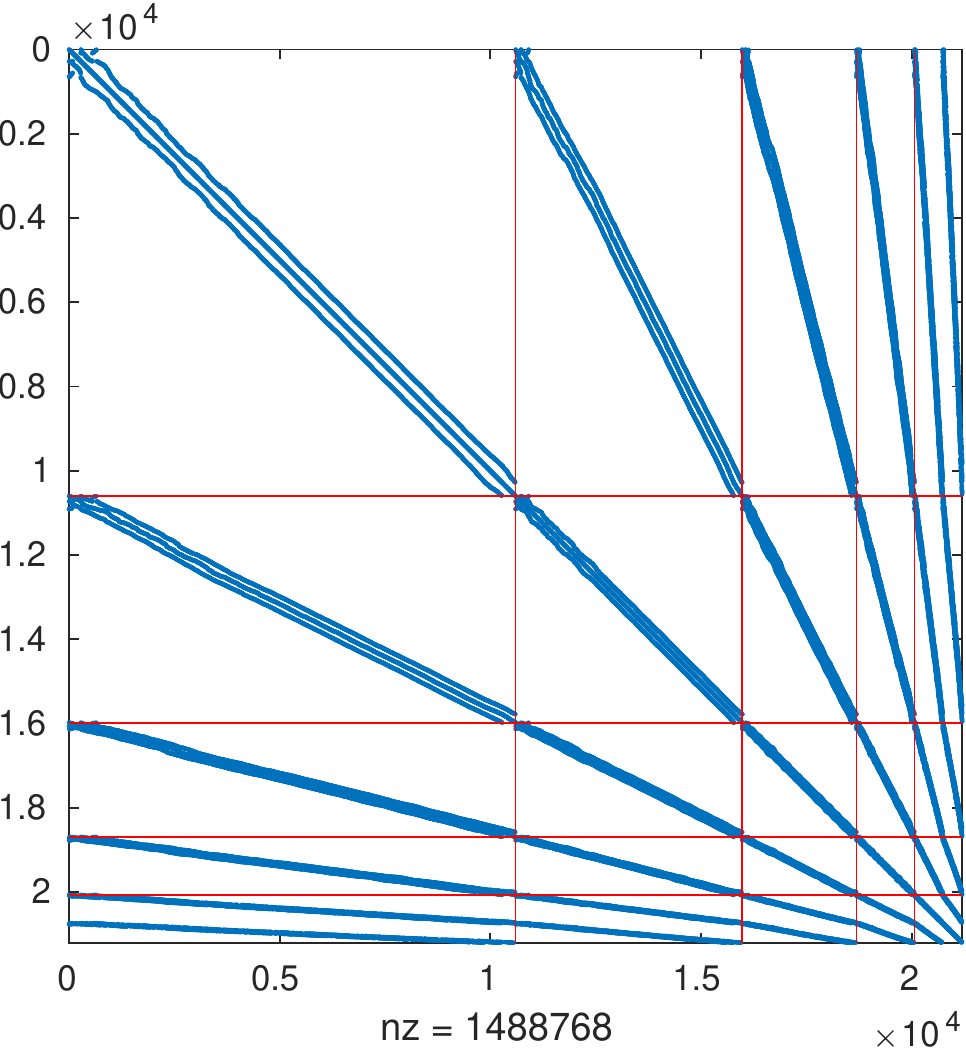}
\includegraphics[height=0.4\textwidth]{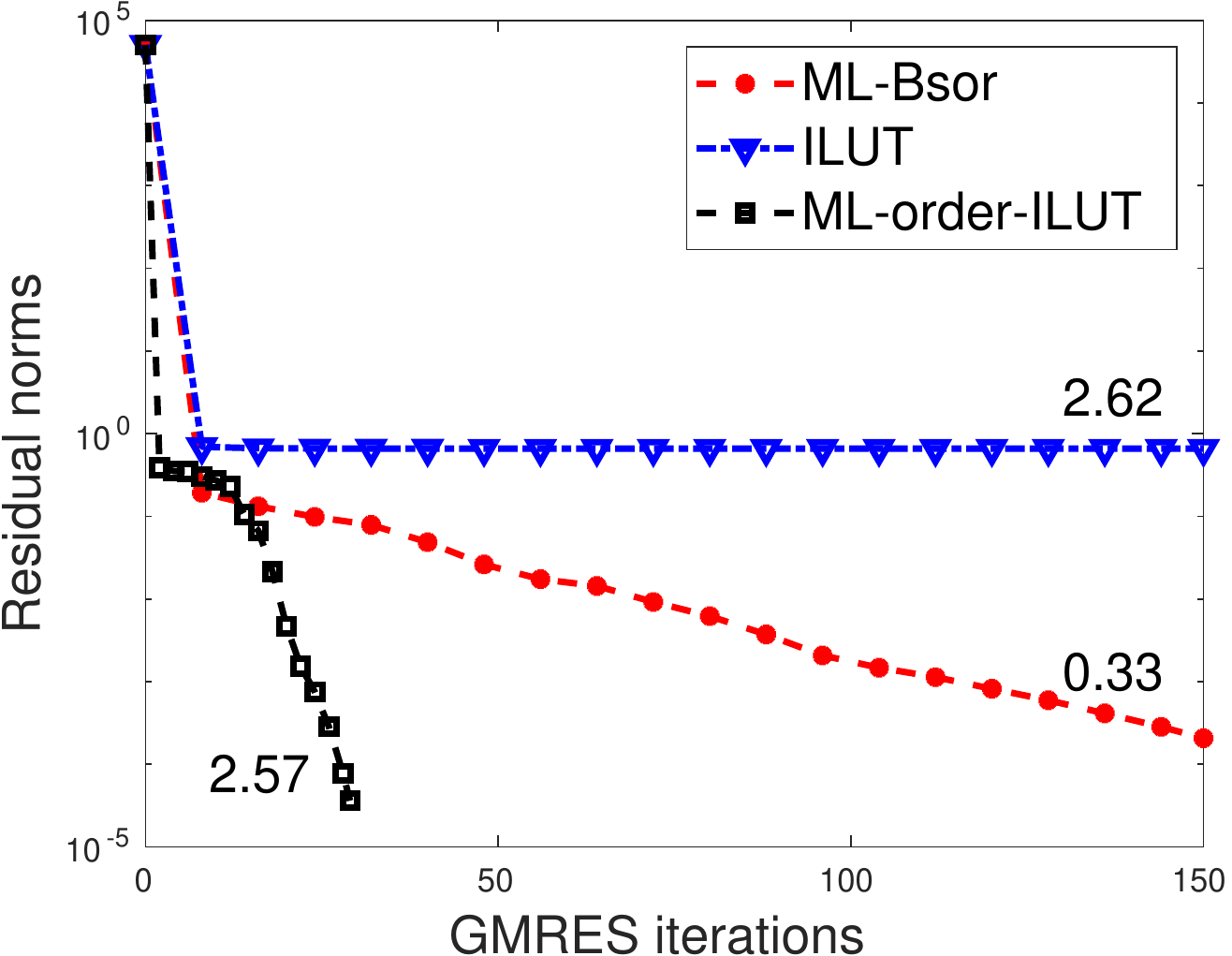}
\end{center}
\caption{Left: The matrix Raefsky3 after the reordering obtained from four
  levels of coarsening.
  Right:
  Performance of various coarsening based preconditioners for solving a linear
  system with the matrix.
  \label{fig:raefsky3}}
\end{figure} 

Here, we show an example on the matrix `Raefsky3', which is of size
21,200 and has 1,488,768 nonzero elements. It comes from a fluid
structure interaction turbulence problem and can be obtained from the
suite-sparse collection\footnote{\url{https://sparse.tamu.edu/}}.
Figure~\ref{fig:raefsky3} (left) shows the pattern of the reordered
matrix according to the coarsening strategy described above, using
four levels of coarsening.  The original matrix is not shown but, as
expected, its pattern is very similar to that of the (1,1) block of
the reordered matrix, and has roughly twice the size.

For this particular matrix, standard ILU-based strategies perform
poorly. Using matlab, we applied GMRES(50) to the system,
preconditioned with ILU (`crout' version) with a drop tolerance of
0.01.  The resulting iterates stagnate as indicated by the top curve
in the right side of Figure~\ref{fig:raefsky3}. The number 2.62 is the
`fill-factor,' which is the ratio of the number of nonzero elements of
the LU factors over the the number of nonzero elements of the original
matrix. When comparing preconditioners of this type, we strive to
ensure that these fill-factors are about the same for the
preconditioners being compared. Next we perform an ILU factorization
(`crout' version again) on the reordered system using the coarsening
described above. In order to achieve a fill-factor similar to that of
the standard factorization we lowered the drop-tolerance to 0.0008.
The new method converged in 29 iterations with a fill factor of 2.57.
We also tested a more traditional preconditioner based on block
Gauss-Seidel, exploiting the block structure shown on the left side
of Figure~\ref{fig:raefsky3}. Each block Gauss-Seidel step requires solving
a system with the diagonal blocks of the reordered matrix. These systems are
approximately solved using a simple ILU(0) (`nofill')
factorization. Note that the fill-factor here is very low (0.33). Each
preconditioning step consists of 10 Gauss-Seidel iterations.  As is
shown, this also converges, although more slowly.

%%%%%%%%%%%%%%%%%%%%%%%%%%%%%%%%%%%%%%%%%%%%%%%%%%%%%%%%%%%%
\subsection{Coarsening approach: Pairwise aggregation}\label{sec:HEM}
\label{sec:aggreg}
The broad class of `pairwise aggregation' techniques,
e.g., \cite{vanvek-1996, vanvek-2001, Notay-2008, Notay-2010,
Chen-book, MG-book}, is a strategy that seeks to simply coalesce two
adjacent nodes in a graph into a single node, based on some measure of
nearness or similarity.  The technique is based on edge
collapsing \cite{YHu}, which is a well known method in the multilevel
graph partitioning literature. In this method, the collapsing edges
are usually selected using the \textit{maximal matching} method.
A \textit{matching} of a graph $G=\left( V,E \right)$ is a set of
edges $\widetilde{E}$, $\widetilde{E}\subseteq E$, such that no two
edges in $\widetilde{E}$ have a node in common.  A maximal matching is
a matching that cannot be augmented by additional edges to obtain a
larger matching in the sense of inclusion.  Coarsening schemes based
on \emph{edge matching} have been in use in the AMG literature for
decades~\cite{Ruge-Stuben-AMG}. For each node $i$, a coarsening
algorithm starts from building a set $S_i$ of nodes that are `strongly
connected' to $i$ by using some measure of connection strength. The
graph nodes are traversed in a certain order of preference and the
next unmarked node in this order, say $j$, is selected as
a \emph{coarse} node.  The priority measure of the traversal is
updated after each insertion of a coarse node.  There are a number of
ways to find a maximal matching for coarsening a graph.

The \textit{heavy-edge matching} (HEM) approach,
e.g., \cite{Karypis95}, is a greedy matching algorithm that works with
the weight matrix $A$ of the graph.  It simply matches a node $i$ with
its largest off-diagonal neighbor $j_{max}$; i.e., we have $ |
a_{ij_{max}} | = \max_{j \in adj (i), j\ne i} | a_{ij}|$, where
$adj(i)$ denotes the adjacency (or nearest-neighbor) set of node $i$.
When selecting the largest neighboring entry, ties are broken
arbitrarily. If $j_{max}$ is already matched with some node $k\neq i$
seen before, i.e., $p(k) = j_{max}$, then node $i$ is left unmatched
and considered as a singleton.  Otherwise, we match $i$ with $j$ and
the result is $p(i)=j_{max}$.

    \begin{algorithm}[ht]
      \caption{Heavy Edge Matching (HEM)}
      \begin{algorithmic}[1]
        \State {\bfseries Input:} Weighted graph $G=(V,E,A)$
        \State {\bfseries Output:} Coarse nodes; $Prnt$ list
        \State {\bfseries Init:} $Prnt(i)=0$ $\forall i\in V$; $new=0$
        \For{$\max$ {\bfseries to} $\min$ edge $(i,j)$}
        \If{$Prnt(i)==0$, $Prnt(j)==0$}
        \State $new = new + 1$
        \State $Prnt(i)=Prnt(j)=new$
        \EndIf
        \EndFor
        \For{Node $v$ with $Prnt(v)==0$}
        \If{$v$ has no neighbor}
        \State $new=new+1$; $Prnt(v)=new$
        \Else
        \State $Prnt(v)=Prnt(j)$ where $j=\argmax_i(a_{iv})$
        \EndIf
        \EndFor
      \end{algorithmic}
      \label{alg:HEM}
    \end{algorithm}

\begin{figure}[ht]
\vskip 0.2in
\begin{center}
\centerline{\includegraphics[width=0.6\textwidth]{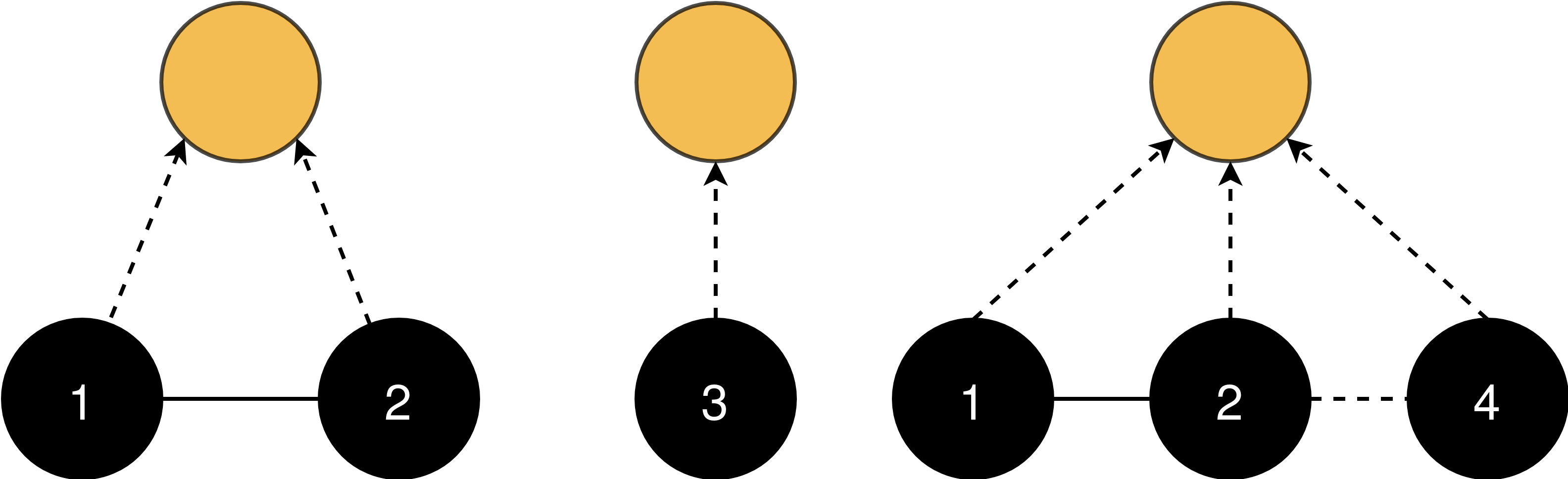}}
\caption{The coarsening process. Original and coarse nodes are colored in black and yellow, respectively. Dashed arrow indicates parent-child relationship. Solid line represents heaviest-weighted edge. \textit{Left:} A coarse node is created from two adjacent nodes $1$ and $2$. \textit{Middle}: A coarse node is created from a true singleton node $3$. \textit{Right}: A left-over singleton node $4$ is attached to a coarse, nearest neighbor node $2$.}
\label{fig:singleton}
\end{center}
\vskip -0.2in
\end{figure}

A version of HEM is shown in Algorithm~\ref{alg:HEM}, modified
from~\cite{HEM}. The algorithm proceeds by exploiting a greedy
approach. It scans all edges $(i,j)$ in decreasing value of their
weight $a_{ij}$. If neither $i$ nor $j$ has defined parents, it
creates a new coarse node labeled $new$ and sets the parents of $i$
and $j$ to be $new$. After the loop is completed, there will be
singletons; i.e., nodes that have not been assigned a parent in the
loop. As shown in lines 10--16, a `singleton' node is either added as
a coarse node, if it is disconnected (`real singleton'), or it is
lumped as a child of an already generated coarse node (`left-over
singleton'). Figure \ref{fig:singleton} gives an illustration of this
step.

%%%%%%%%%%%%%%%%%%%%%%%%%%%%%%%%%%%%%%%%%%%%%%%%%%%%%%%%%%%%
\subsection{Coarsening approach:  Independent sets} \label{sec:ISC}
It is also possible to exploit independent sets (see, e.g.,
\cite{HRFangEtAlcikm1}) for coarsening graphs. Recall that an
independent set $\calS$ is a subset of $V$ that consists of nodes that
are not adjacent to each other; i.e., no pair $(v,w)$ where $v, w \in
\calS$ is linked by an edge.  An independent set $\calS$ is maximal if
no (strict) superset of $\calS$ forms another independent set. One can
use the nodes of a carefully selected independent set to form a coarse
graph; i.e., we can define $V_c = \calS$.  Then, it is relatively easy
to determine the edges and weights between these nodes by using
information from the original graph. For example, in
\cite{HRFangEtAlcikm1} an edge is inserted between $v$ and $w$ of
$V_c$ if there is at least one node $y$ in $V \backslash V_c$ such
that $(v,y) \in E$ and $(w,y) \in E$.  This will produce the edge set
$E_c$ needed to form the coarse graph $G_c$.

Let $L$ be the graph Laplacian, reordered such that the $n_c$ nodes of
$V_c$ are listed first. Then $L$ will have the following structure,
where $D_c \ \in \ \RR^{n_c\times n_c}$~:
\eq{eq:is0}
L = \begin{pmatrix}
  D_c & -F \\ -F^T & B \end{pmatrix}.
\en

Since the nodes associated with the (1,1) block in the above matrix
form an independent set, it is clear that the matrix $D_c$ is
diagonal.  Coarsening by independent sets will consist of taking the
matrix $D_c$ and adding off diagonal elements to it to obtain the
adjacency matrix $A_c$.  An important observation to be made here is
that the edges added by the independent set coarsening are those
obtained from the Schur complement with respect to $B$, \emph{where
$B$ is replaced by a diagonal matrix}. Let us assume that $B$ is
replaced by a matrix $D_f$. The resulting Schur complement is
\eq{eq:Sc}
  S_c = D_c - F D_f\inv F^T.
\en
The nonzero pattern of $S_c$ is the same as that of $FF^T$, which is
in turn the sum of the patterns of $f_j f_j^T$ where $f_j$ is the
$j$-th column of $F$.  If $f_j $ has nonzero entries in locations $i$
and $k$, then we will have nonzero entries in the positions $(i,i),
(i,k), (k,i)$ and $(k,k)$ of $S_c$.  Next, we will define $D_f$ more
specifically. Let $D_f$ be the diagonal of row-sums of $F^T$ and
assume for now that these are all nonzero.  This is the same diagonal
as the one used for the graph Laplacian except that the summation
ignores the entries $a_{ij}$ when $i$ and $j$ are both in $
V\backslash V_c$. In matlab notation: $D_f = \diag (\diag (F^T
\ones))$. In this situation, $S_c$ becomes a Laplacian.

\begin{proposition}
  Let $B$ be replaced by $D_f$, defined as the diagonal of the
  row-sums of $F^T$.  Then $D_f$ is invertible. Let $L_c = D_c - F
  D_f\inv F^T$. Then the graph of $L_c$ is $G_c$, the graph of the
  independent set coarsening of $G$. In addition, $L_c$ is a graph
  Laplacian; specifically, it is the graph Laplacian of $G_c$.
\end{proposition}

\begin{proof}
  Because the independent set is maximal, we cannot have a zero
  diagonal element in $D_f$. Indeed, if the opposite was true, then
  one row, say row $k$, of $F^T$ would be zero. This would mean that
  we could add node $k$ to the independent set, because it is not
  coupled with any element of $\calS$. The result would be another
  independent set that includes $\calS$, contradicting maximality.

  It was shown above that the adjacency graph of $F F^T$, which is the
  same as that of $L_c$, is exactly $G_c$.
  It is left to show that $L_c$ is a Laplacian.  Since $D_f$ and $F$
  have nonnegative entries, the off-diagonal of $L_c $ are clearly
  negative.  Next, note that $F^T \ones = D_f \ones $ and hence
  \[
  (D_c - F D_f\inv F^T)\ones = D_c\ones - F D_f\inv D_f\ones =
  (D_c - F ) \ones = 0.
  \]
  Thus, $L_c$ is indeed a Laplacian. 
\end{proof}

%%%%%%%%%%%%%%%%%%%%%%%%%%%%%%%%%%%%%%%%%%%%%%%%%%%%%%%%%%%%
\subsection{Coarsening approach:  Algebraic distance}\label{sec:ALG}
Researchers in AMG methods defined a notion of `algebraic distance'
between nodes based on relaxation procedures.  This notion is
motivated by the bootstrap AMG (BAMG) method \cite{brandt-review-01}
for solving linear systems. AMG creates a coarse problem by trying to
exploit some rules of `closeness' between variables. In BAMG, this
notion of closeness is defined from running a few steps of
Gauss-Seidel relaxations, starting with some random initial guess for
solving the related homogeneous system $Ax= 0$.  The speed of
convergence of the iterate determines the closeness between
variables. This is exploited to aggregate the unknowns and define
restriction and interpolation operators \cite{Safro-2011}.  In
\cite{ChenSafro-2011} this general idea was extended for use on graph
Laplacians. In the referenced paper, Gauss-Seidel is replaced by
Jacobi overrelaxation.

\begin{algorithm}[h]
  \caption{Algebraic distances for graphs}
  \begin{algorithmic}[1]
    \State {\bfseries Input:} Parameter $\omega$, weighted graph $G=(V,E,A)$,
    initial vector $x\up{0}$, number of steps $k$.
    \State {\bfseries Output:} Distances $s_{ij} \up{k}$ for each pair $(i,j)$.
    \For{$j=1,2,\cdots,k$}
    \State $ x\up{j} = (1-\omega) x\up{j-1} +  \omega D\inv A x\up{j-1} $
    \EndFor
    \State Set: $ s_{ij}\up{k} = |  x_i\up{k} - x_j\up{k}|  \,\, \forall i,j$
  \end{algorithmic}
  \label{alg:AlgD}
\end{algorithm}

Algorithm~\ref{alg:AlgD} shows how these distances are calculated.
They depend on two parameters: the over relaxation parameter $\omega$
and the number of steps, $k$. It can be shown that the distances
$s_{ij}\up{k}$ converge to zero~\cite{ChenSafro-2011} as $k\rightarrow
\infty$.  However, it was argued in \cite{ChenSafro-2011} that the
speed of decay of $s_{ij}\up{k}$ is an indicator of relative strength
of connection between $i$ and $j$. In other words, the important
measure is the magnitude -- in relative terms -- of $s_{ij}\up{k}$ for
different $(i, j)$ pairs.

Note that the iteration is of the form $x\up{j} = H x\up{j-1}$, where
$H$ is the iteration matrix
\[ H = (1-\omega) I + \omega D\inv A = I - \omega (I-D\inv A) . \]
Because $D-A$ is a graph Laplacian, the largest eigenvalue of $H$ is
$\lambda_1 =1$.  It is then suggested to scale these scalars by
$\lambda_2^k$, where $\lambda_2$ is the second largest eigenvalue in
modulus.  In general, it is sufficient to iterate for a few steps and stop at a step
$k$ when one observes the scaled quantities start to settle.
  
As can be seen, a coarsening method based on algebraic distances is
rather different from the previous two methods. Instead of working on
the graph directly we now use our intuition on the iteration matrix to
extract intuitive information on what may be termed a relative
distance between variables.  If two variables are close with respect
to this distance, they may be aggregated or merged.
  
Ultimately, as was shown in \cite{ChenSafro-2011}, what is important
is the decay of the component of the vector $x\up{k}-x\up{k-1}$ in the
second eigenvector. This distance between two vectors is indeed
dominated by the component in the second eigenvector.

This brings up the question as to whether or not we can directly
examine spectral information and infer from it a notion of distance on
nodes.  Spectral graph coarsening addresses this and will be examined
in Section~\ref{sec:spectral}.

%%%%%%%%%%%%%%%%%%%%%%%%%%%%%%%%%%%%%%%%%%%%%%%%%%%%%%%%%%%%
\subsection{Techniques related to coarsening}\label{sec:coarse.related}
Graph coarsening is a \emph{graph reduction} technique, in the sense
that it aims at reducing the size of the original graph while
attempting to preserve its properties. There exist a number of other
techniques in the same category. These include \emph{graph
summarization} \cite{navlakha2008graph,liu2018graph}, \emph{graph
compression} \cite{fan2012query}, and \emph{graph sketching}
~\cite{krauthgamer2020sketching}.  These are more common in machine
learning and the tasks they address are specific to the underlying
applications.  In the following we discuss methods that are more akin
to standard coarsening methods.
  
\subsubsection{Graph reduction: Kron} \label{sec:kron} 
The Kron reduction of networks was proposed back in 1939
\cite{kron39}, as a means to obtain lower dimensional electrically
equivalent circuits in circuit theory. Its popularity gained momentum
across fields after the appearance of a thorough analysis of the
method in \cite{Kron-paper13}.  The method starts with a weighted
graph $G = (V, E, A)$ and the associated graph Laplacian $L$, along
with a set $V_1$ of nodes which is a strict subset of $V$. Such a
subset can be obtained by downsampling, for example, although in the
original application of circuits it is a set of nodes at the boundary
of the circuit.  The method essentially defines a coarse graph from
the Schur complement of the original adjacency graph with respect to
this downsampled set.

The goal is to form a reduced graph on $V_1$. This is viewed from
the angle of Laplacians.  If we order the nodes of $V_1$ first,
followed by those of the complement set to $V_1$ in $V$, the Laplacian
can be written in block form as follows:
\eq{eq:LapBlock} L
= \begin{bmatrix} L_{11} & L_{12} \\ L_{12}^T & L_{22} \end{bmatrix} .
\en
The Kron reduction of $L$ is defined as the Schur complement of the
original Laplacian relative to $L_{22}$; i.e.,
\eq{eq:schur} L(V_1) =
L_{11} - L_{12} L_{22}\inv L_{12}^T .
\en

\begin{figure}
\includegraphics[height=0.4\textwidth]{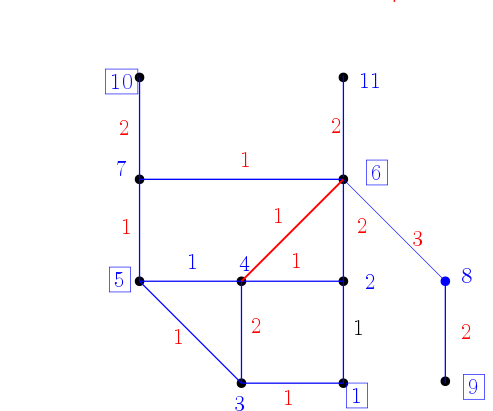} 
\hfill\raisebox{30pt}{\includegraphics[height=0.2\textwidth]{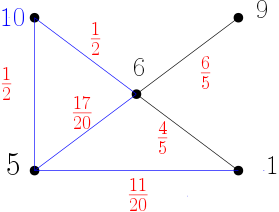}} \hfill
\rule{0pt}{2pt} 
\caption{A small graph (left) and its Kron reduction (right).
  The reduced set of nodes is $V_1= \{ 1, 5, 6, 9, 10\}$. These node
  labels are  kept in the Kron reduction. Numbers on the middle of the
edges are the weights.}  
\label{fig:kronex} 
\end{figure}

This turns out to be a proper graph Laplacian as was proved in
\cite{Kron-paper13}, along with a few other properties.  We can
therefore associate a set of weights $A_{ij}\up{1}$ for the reduced
graph, defined from $L(V_1)$:
\[
a_{ij}\up{1} = \left\{
\begin{array}{cl} - [L(V_1)]_{ij} & \mbox{if } i\ne j, \\
  0         & \mbox{otherwise}.
\end{array} \right.
\]
An example is shown in Figure~\ref{fig:kronex}.

A multiscale version of the Kron reduction, called the \emph{pyramid
transform}, was proposed in \cite{MSkron}, specifically for
applications that involve signal processing on graphs
\cite{shuman-survey}.  It was developed as a multiscale (i.e.,
`multilevel' in the scientific computing jargon) extension of a
similar scheme invented in the late 1980s for image
processing~\cite{BurtAdelson87}.  The extension is from regular data
(discrete time signals, images) to irregular data (graphs, networks)
as well as from one level to multiple levels.

An original feature of the paper \cite{MSkron} is the use of spectral
information for coarsening the graph. Specifically, a departure from
traditional coarsening methods such as those described in
Sections~\ref{sec:HEM} and~\ref{sec:ISC} is that the separation into
coarse and fine nodes is obtained from the `polarity' (i.e., the sign
of the entries) of the eigenvector associated with the largest
eigenvalue of the graph Laplacian.  The motivation of the authors is a
theorem by Roth \cite{Roth89}, which deals with bipartite graphs. The
idea of exploiting spectral information was exploited earlier by
Aspvall and Gilbert \cite{AspvallGilbert84} for the problem of graph
coloring, an important ingredient of many linear algebra techniques.
Another original feature of the paper \cite{MSkron} is the use of
spectral methods for sparsifying the Schur complement. As was
mentioned earlier, the Schur complement will typically be dense, if
not full in most situations.  The authors invoke `sparsification' to
reduce the number of edges. We will cover sparsification in
Section~\ref{sec:sparsify}.

\paragraph{Example}
As an example, we return to the illustration of
Figure~\ref{fig:kronex}.  Using normalized Laplacians, we find that
the largest eigenvector separates the graph in two parts according to
its polarity, namely $V_1 = \{ 1, 5, 6, 9, 10 \} $ and $V_2=\{2, 3, 4,
7, 8, 11\}$.  Thus, it is able to discover $V_1$, the rather natural
independent set we selected earlier.

One important question that can be asked is \emph{why resort to the
Schur complement as a means of graph reduction?}  A number of
properties regarding the Kron reduction were established in
\cite{Kron-paper13} to provide justifications.  Prominent among these
is the fact that the resistance distance
\cite{ellens2011effective} between nodes of the coarse graph are
preserved. The resistance distance involves the pseudo-inverse of the
Laplacian.

\subsubsection{Relationship between Kron reduction and independent set coarsening} 
\begin{figure}[h]
  \centerline{\includegraphics[width=0.6\textwidth]{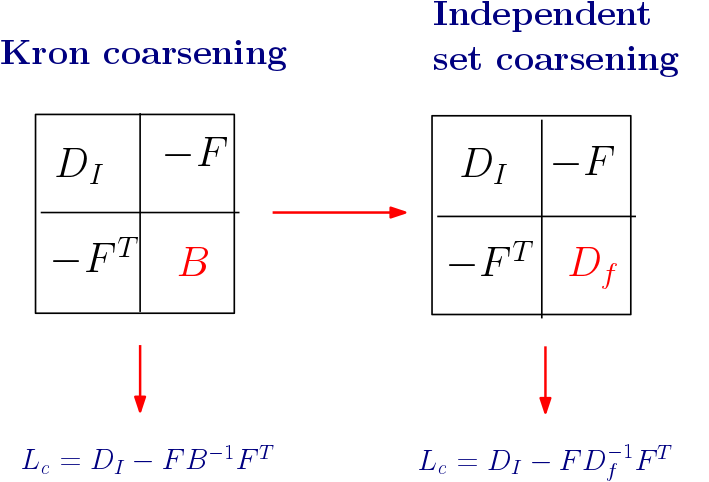}}
  \caption{Two ways of using independent sets for coarsening.\label{fig:isets}}
\end{figure}

Instead of invoking independent sets, the Kron reduction, as it is
used in \cite{MSkron}, `downsamples' nodes by means of spectral
information.  While these samples may form an independent set, this is
not guaranteed.  Just like independent set coarsening, the coarse
graph is built from the Schur complement associated with the
complement of the independent set (so-called \emph{node cover}).
These two ways of coarsening are illustrated in
Figure~\ref{fig:isets}.

Unlike independent set coarsening, the matrix $L_{22}$ involved the
Schur complement of \nref{eq:schur} is not approximated by a diagonal
before inversion, in effort to reduce the fill-ins introduced.
Instead, spectral `sparsification' is invoked.  In the survey paper
\cite{Kron-paper13}, the set $V_1$ represents a set of `boundary
points' in an electrical network.  Improved sparsity is achieved by
other means than those employed in Section \ref{sec:ISC}, specifically
by eliminating a selection of internal nodes instead of all of them,
in the Gaussian elimination process.

\subsubsection{Graph sparsification}\label{sec:sparsify}
Graph sparsification methods aim at finding a sparse approximation of
the original graph by trimming out edges from the graph
\cite{spielmanTeng11,spielmanSriv11}.  Here, the number of nodes
remains the same but the number of edges, $|E|$, can be much lower than
in the original graph.  This is motivated by the argument that in some
applications, the graphs that model the data tend to be rather dense
and that many edges can be removed without negatively impacting the
result of methods that exploit these graphs; see, e.g.,
\cite{kapralov2014single}.

Different measures of closeness between the sparsified and the
original graph have been proposed for this purpose, resulting in a
wide range of strategies such as spanners \cite{Althofer1993}, cut
sparsifiers \cite{cutsparse}, and spectral sparsifiers
\cite{spielmanTeng11}.  Graph sparsification methods can be beneficial
when dealing with high density graphs and come with rigorous
theoretical guarantees \cite{batson2013Spectral}; see also
\cite{BenczurKarger96}.

We already mentioned at the end of Section~\ref{sec:kron} one specific
use of graph sparsifiers in the context of multilevel graph
coarsening. To sparsify the successive Schur complement obtained by
the multilevel scheme, the authors of \cite{MSkron} resorted to a
spectral sparsifier developed in \cite{spielmanSriv11}.  If
$\widetilde G$ is the sparsified version of $G$, and if $\widetilde L$ and
$L$ are their respective Laplacians, the main goal of spectral
sparsifiers is to preserve the quadratic form associated with
the Laplacians. The graph $\widetilde G$ is said to be a
$\sigma$-spectral approximation of $G$ if for all $x \ \in \ \RR^n$,
\eq{eq:sig-app}
\frac{1}{\sigma} x^T \widetilde{L} x \le x^T L x \le
\sigma \ x^T \widetilde L x .
\en
A trivial observation for $\sigma$-similar ($\sigma>1$) graphs is that
their Rayleigh quotients
\[  \mu(x) = \frac{(L x,x)}{(x,x)} \ ,
\qquad
\widetilde \mu(x) = \frac{(\widetilde L x,x)}{(x,x)}
\]
for the same nonzero vector $x$  satisfy the double inequality:
\eq{eq:sig-mu}
\frac{1}{\sigma} \ \widetilde \mu(x) \le \mu(x) \le   \sigma \ \widetilde \mu(x) .
\en
Thus, these Rayleigh quotients are, in relative terms, within a factor
of $\sigma-1$ of each other:
\[
\left| \frac{\widetilde \mu (x) - \mu (x)}{\widetilde \mu (x)} \right| \le \sigma - 1 .
\]

This has an impact on eigenvalues.  If $\sigma $ is close to
one, then clearly the eigenvalues of $L$ and $\widetilde L$ will be
close to each other, thanks to the Courant--Fisher min-max
characterization of eigenvalues~\cite{GVL-book}.  In what follows,
$S_k$ represents a generic $k$-dimensional subspace of $\RR^n$ and
eigenvalues are sorted decreasingly. In this situation, the theorem
states that the $k$-th eigenvalue of the Laplacian $L$ satisfies:
\eq{eq:courant}
\lambda_k = \max_{\text{dim} (S_k)=k} \quad \min_{0
  \ne x \in S_k} \mu(x) .
\en
The above maximum is achieved by a set, denote by $S_*$ (which is just
the linear span of the set of eigenvectors $u_1, \cdots, u_k$).  Then
\[
  \lambda_k = \min_{0 \ne x  \in S_*} \ \mu(x)
  \,\,\le\,\, \min_{0 \ne x  \in S_*}  \ \sigma  \widetilde \mu(x) 
  \,\,\le\,\, \sigma \max_{S_k} \min_{0 \ne x  \in S_k}  \ \widetilde \mu(x)
  \,\,\le\,\, \sigma \ \widetilde \lambda_k . 
\]
The exact same relation as \nref{eq:sig-app} holds if $L$ and
$\widetilde L$ are interchanged. Therefore, the above relation also
holds if $\lambda$ and $\widetilde \lambda$ are interchanged, which
leads to the following double inequality, valid for $k=1, 2, \cdots,
n$:
\eq{eq:compareig}
\frac{\widetilde \lambda_k}{\sigma}    \le  \lambda_k \le \sigma \widetilde \lambda_k . 
\en

%%\hl{REF? The above simple result may already exist in the literature?}

It is also interesting to note a link with preconditioning techniques.
When solving symmetric positive definite linear systems of equations,
it is common to approximate the original matrix $A$ by a
preconditioner which we denote here by $\widetilde A$. Two desirable
properties that must be satisfied by a preconditioner $\widetilde A$
are that (i) it is inexpensive to apply $\widetilde A\inv $ to a
vector; and (ii) the condition number of $\widetilde A \inv A$ is
(much) smaller than that of $A$. The second condition translates into
the condition that $ (Ax,x)/(\widetilde A x,x) $ be small. If we
assume that
\eq{eq:condPrec}
\frac{1}{\sigma} \le \frac{(Ax,x)}{(\widetilde A x, x)} \le \sigma,
\en
then the condition number of the preconditioned matrix, which is the
ratio of the largest to the smallest eigenvalues of $\widetilde A\inv
A$, will be bounded by $\sigma^2$.

\subsubsection{Graph partitioning} 
The main goal here is to put in contrast the problem of coarsening
with that of graph partitioning. To his end, a brief background is
needed.  In spectral graph partitioning
\cite{Fiedler73,BarSim,PotSimLiou}, the important equality
\nref{eq:LaplProp} satisfied by any Laplacian $L$ is exploited.  If
$x$ is a vector of entries $+1$ or $-1$, encoding membership of node
$i$ to one of two subgraphs, then the value of $x^T L x$ is equal to 4
times the number of edge cuts between the two graphs with this 2-way
partitioning.  We could try to find an optimal 2-way partitioning by
minimizing the number of edge cuts, i.e., by minimizing $x^T Lx$
subject to the condition that the two subgraphs are of equal size,
i.e., subject to $\ones^T x = 0$. Since this optimization problem is
hard to solve, it is common to `relax' it by replacing the conditions
$x \in \{-1,1\}^n, x^T \ones=0$ with $x \in \RR^n, \|x\| = 1, x^T
\ones=0$.  This leads to the definition of the Fiedler vector, which
is the second smallest eigenvector of the Laplacian.  Recall that the
smallest eigenvalue of the graph Laplacian is zero and that when the
graph is connected, this eigenvalue is simple and the vector $\ones$
is a corresponding eigenvector.

It is interesting to note the similarity between spectral graph
partitioning and Kron reduction. In both cases, the polarity of an
eigenvector is used to partition the graph in two subgraphs. In the
case of Kron reduction, it is the vector associated with the largest
eigenvalue that defines the partitions; and one of these partitions is
selected as the `coarse', or the `downsampled' set, according to the
terminology in \cite{MSkron}.  For graph partitioning, what is done
instead is to use the eigenvector at the other end, the one next to
the smallest, since the smallest is a constant vector.  If we
reformulate the problem back in terms of assignment labels of $\pm 1$,
then this would lead to the interpretation that in one case, we try to
minimize edge cuts (partitioning) and in the other, we try to maximize
them (Kron reduction).

\begin{figure}[h]
  \begin{center}  \includegraphics[height=0.45\textwidth]{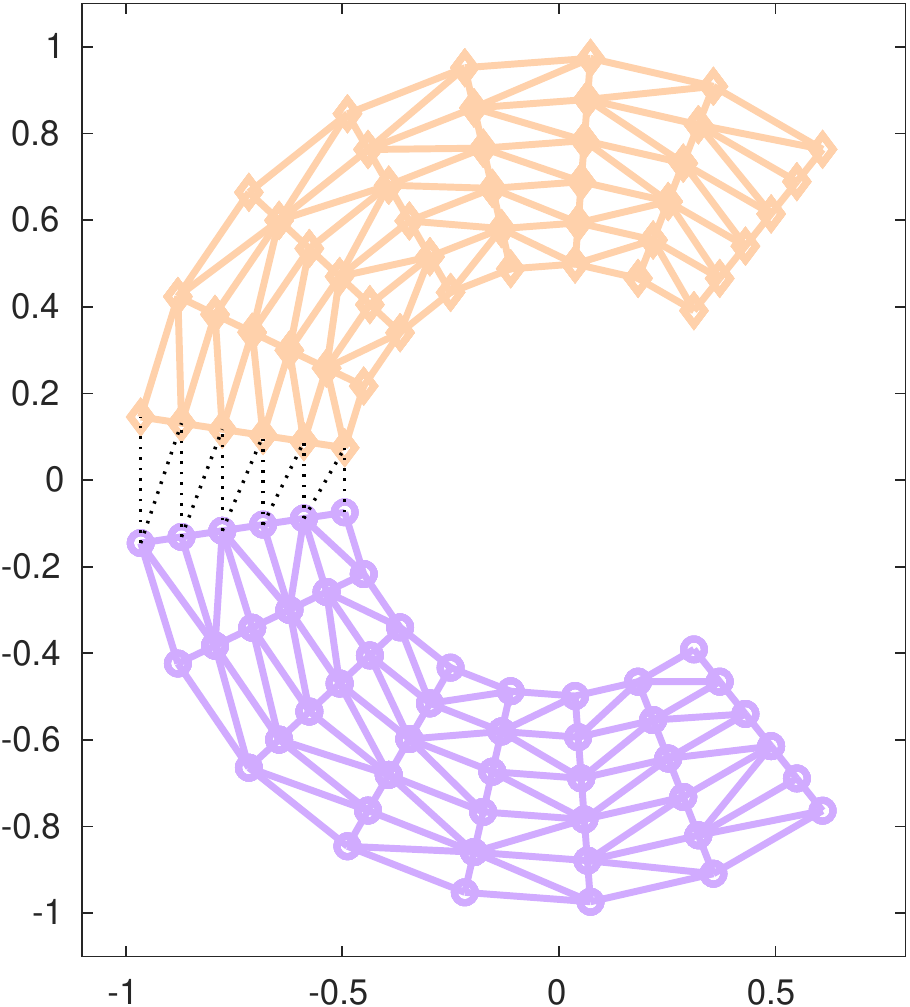}
    \includegraphics[height=0.45\textwidth]{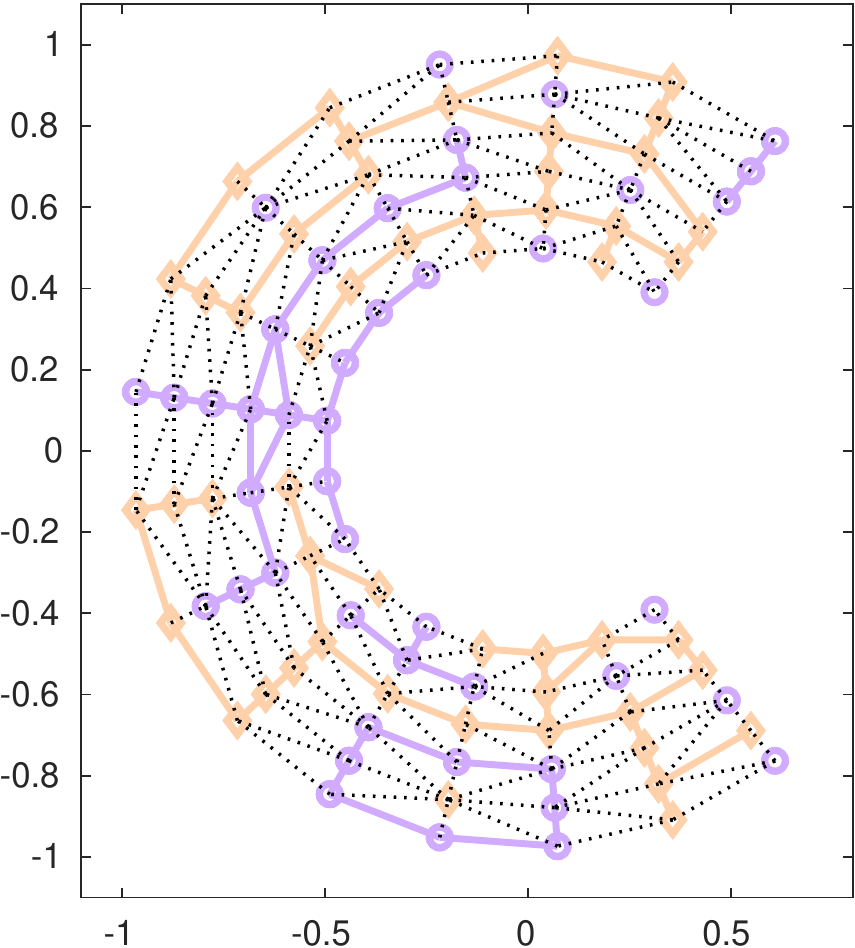}
  \end{center}
  \caption{Subgraphs from eigenvectors of the (normalized)
    Laplacian.  Left side: using second smallest eigenvector;
    Right side: using largest eigenvector. The two sets are colored differently
    and edge-cuts are shown as dotted lines.}\label{fig:subg}
\end{figure}

An illustration is shown in Figure~\ref{fig:subg} with a small finite
element mesh.  As can be seen, using the second smallest eigenvector
tends to color the graph in such a way that nearest neighbors of a
node are mostly of the same color, while using the largest eigenvector
tends to color the graph in such a way that nearest neighbors of a
node are mostly of a different color.  Another way to look at this is
that using the second smallest eigenvector gives domains that tend to
be `fat' whereas the largest eigenvector gives domains that tend to be
`thin', like unions of lines separating each other. Another fact shown
by this simple example is that neither of the two sets obtained is
close to being an independent set.

%%\hl{This simple result may already exist in the literature?}

The following property is straightforward to prove.
\begin{proposition}
Assume that the graph has no isolated node and that the components
$\xi_1, \xi_2, \cdots, \xi_n$ of the largest eigenvector $u_1$ are
nonzero.  Let $V_+$ and $V_-$ be the two subgraphs obtained from the
polarities of the largest eigenvector. Then each node of $V_+$
(resp. $V_-$) must have at least one adjacent node from $V_-$
(resp. $V_+$).
\end{proposition}
\begin{proof}
  The $i$-th row of the relation $L u_1 = \lambda_1 u_1$ yields: (recall definition
  \nref{eq:Glap}) 
\[ d_i \xi_i - \sum_{j\in \ N(i) } a_{ij} \xi_j  = \lambda_1 \xi_i  \quad
\rightarrow \quad  (\lambda_1 - d_i)  \xi_i = - \sum_{j\in \ N(i) } a_{ij} \xi_j  . \]
Note that $(\lambda_1 - d_i) > 0$ (due to assumption). Then, if $\xi_i
\ne 0$ (left side) then at least one of the $\xi_j$'s, $j \ne i$
(right side) must be of the opposite sign.
\end{proof} 

%%\hl{   To Add:   Coarsening helps partitioning.. referred to below.}

%%%%%%%%%%%%%%%%%%%%%%%%%%%%%%%%%%%%%%%%%%%%%%%%%%%%%%%%%%%%
\section{Graph coarsening in machine learning}\label{sec:ML}
In this section, we discuss existing methods related to graph
coarsening in machine learning and discuss how they are employed in
typical applications.  We begin by defining the types of problems
encountered in machine learning and the related  terminology.
The terms `graphs' and `networks' are often used interchangeably in the
literature -- although the term networks is often employed
specifically for certain types of applications, whereas graphs are
more general.  For example, `social networks' are common for
applications that model friendship or co-authorship, while one speaks
of graphs when modeling chemical compounds. What muddles the
terminology is the term `graph neural networks,' which are neural networks (a
type of machine learning model) for graphs. Furthermore, there are
`protein-protein interaction networks,' which are graphs of proteins;
that is, each graph node is a protein, which by itself is a molecular
graph.

Some commonly encountered graph tasks include: node classification
(determine the label of a node in a given graph), link prediction
(predict whether or not an edge exists between a pair of nodes), graph
classification (determine the label of the graph itself), and graph
clustering (group nodes that are most alike together).  One of the key
concepts in a majority of the methods for solving these tasks is
\emph{embedding}, which is vector in  $\RR^d$  used to represent a node
$v\in V$ or indeed the entire graph $G=(V,E)$.

%%%%%%%%%%%%%%%%%%%%%%%%%%%%%%%%%%%%%%%%%%%%%%%%%%%%%%%%%%%%
\subsection{Graph clustering and GraClus}
As was seen earlier, graph coarsening is a basic ingredient of graph
partitioning.  The same graph partitioning tools can be used in
data-related applications, but the requirement of having partitions of
equal size is no longer relevant.  This observation led to the
development of approaches specifically for data applications; see,
e.g., \cite{GraClus}.  The method developed in \cite{GraClus} uses a
simple greedy graph coarsening approach whereby nodes are visited in a
random order. When visiting a node, the algorithm merges it with the
unvisited nearest neighbor that maximizes a certain measure based on
edge and node weights. The visited node and the selected neighbor are
then marked as visited.  The algorithm developed in \cite{GraClus},
which is known as GraClus, uses different tools from those of graph
partitioning. Because it is used for data applications, the refinement
phase exploits a kernel K-means technique instead of the usual
Kernighan--Lin procedure \cite{KerLin}.

%%%%%%%%%%%%%%%%%%%%%%%%%%%%%%%%%%%%%%%%%%%%%%%%%%%%%%%%%%%%
\subsection{Multilevel graph coarsening for node embedding}
\label{sec:embed}
In one form of node embedding, one seeks a mapping $\Phi$ from the
node set $V$ of a graph $G=(V,E)$ to the space $\RR^{n \times d}$
where $n = | V|$ and
\eq{eq:Nembed} \Phi : v \in V \longrightarrow \Phi(v) \ \in \ \RR^d .  \en
In other words, each node is mapped to a vector in $d$-dimensional
space.  Here, the dimension $d$ is usually much smaller than $n$.
Many embedding methods have been developed and used effectively to
solve a variety of problems; see, e.g.,
\cite{LLE-paper,Eigenmaps-paper,Ahmed-al_2013,LINE,GraRep15,Node2Vec16,CommPresEmbed17,HARP-paper18}
and \cite{Goyal2018} for a survey.  The idea of applying coarsening to
obtain embedding for large graphs has been gaining ground in recent
years; see,
\cite{HARP-paper18,MILE-paper18,GraphZoom20,PanayotGraphEmbed21}.

The authors of \cite{HARP-paper18} present a method dubbed
\emph{hierarchical representation learning for networks} (HARP), which
exploits coarsening to facilitate and improve graph embedding.  The
method starts by performing a sequence of $\ell$ graph coarsening
steps to produce graphs $G_1, G_2, \cdots, G_{\ell}$ from the initial
graph $G_0$.  Then, an embedding is performed on the final level to
produce the mapping $\Phi_{\ell}$. This embedding is then propagated
back to the original level by proceeding as follows. Starting from
level $i=\ell$, the mapping $\Phi_i$ is naturally prolongated
(`extrapolated') from level $i$ to level $i-1$ to yield a mapping
$\Phi_{i-1}'$.  An extra step is taken to refine this embedding and
obtain $\Phi_{i-1}$. This specific step is rather reminiscent of the
`post-smoothing' steps invoked in various MG schemes for solving
linear systems.  Post-smoothing consists of a few steps of smoothing
operations, typically a standard relaxation method (e.g.,
Gauss-Seidel), applied after an approximate solution is prolonged from
a coarse level.  The MILE method described in \cite{MILE-paper18} is
rather similar to the HARP approach described above, the main
difference being that the refinement proposed in this method exploits
neural networks.

HARP and MILE are general frameworks that use coarsening to improve
graph embedding. In the remainder of this section, we give an
illustrative example to demonstrate the effectiveness of HARP. We
examine the performance improvement of three widely used graph
embedding algorithms: DeepWalk~\cite{perozzi2014deepwalk},
LINE~\cite{LINE}, and Node2vec~\cite{Node2Vec16}, each combined with
HARP. Furthermore, because HARP is a general framework and we have the
freedom to choose the coarsening method it uses, we examine the impact
of different coarsening methods on the performance improvement.  Three
coarsening methods are tested: HEM (Section~\ref{sec:HEM}), algebraic
distance (Section~\ref{sec:ALG}), and the LESC method to be introduced
in Section~\ref{sec:LESC.s} (it is similar to HEM but uses spectral
information to define the visiting order of nodes).

\begin{figure}[h]
  \begin{center}  \includegraphics[height=0.3\textwidth]{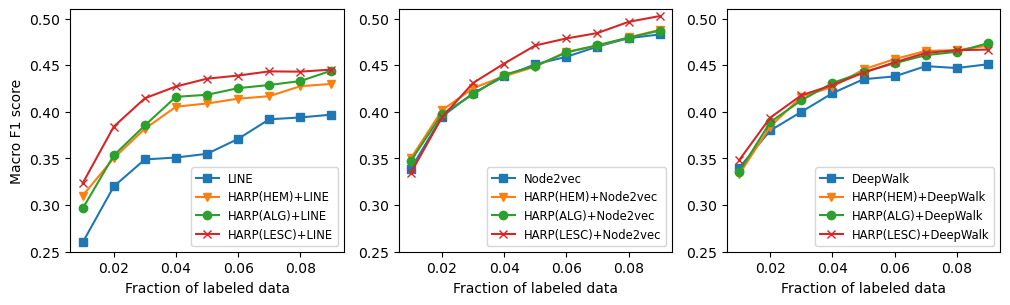}
  \end{center}
  \caption{Classification results on the Citeseer graph. The x-axis shows the portion of nodes used for training. The y-axis shows the macro-average F1 score.  Each reported score is an average over five random repetitions.
  }\label{fig:citeseer}
\end{figure}

We evaluate the HARP framework on a node classification task with the
Citeseer graph~\cite{sen2008collective}.  Citeseer is a citation
graph of computer science publications, consisting of 3.3K nodes and
4.5K edges. The label of each node indicates the subject area of the
paper. The task is to predict the labels.  We first generate the node
embedding for each node using the HARP method. Then, a fraction of the
nodes are randomly sampled to form the training set and the remaining
is used for testing. We train a logistic regression model
\cite{Bishop-book} by using the training data and evaluate the
classification performance on the test data. We use the macro-average
F1 score \cite{Powers08} as the performance metric, which is the mean
of the F1 score for each label category.%
\footnote{An alternative definition, which is less used, is that the
macro-average F1 score is the harmonic mean of the macro-average precision
and the macro-average recall, where the macro-average precision/recall
is the mean of the precision/recall for each label category.}
Figure~\ref{fig:citeseer} shows the score under different training
set sizes.

As can be seen the HARP framework consistently improves all these
embedding methods, especially LINE and DeepWalk. Each coarsening
approach used inside the HARP framework improves the performance to a
different degree, with the LESC approach generally outperforming the
others.

%%%%%%%%%%%%%%%%%%%%%%%%%%%%%%%%%%%%%%%%%%%%%%%%%%%%%%%%%%%%
\subsection{Graph neural networks and graph pooling} \label{sec:GNN}
GNNs have recently attracted a great deal of interest in various
disciplines, including biology~\cite{dobson_distinguishing_2003},
chemistry~\cite{NCI1}, and social networks~\cite{DeepGraphKernels},
where data is modeled by graphs. A major class of GNNs are those of a
convolution style that  generalize lattice convolutions in
convolutional neural networks (CNNs). A standard convolution applies a
filter on a signal; in CNNs, the signal is a 2-dimensional image and
the filter has a very small support---say, a $3\times3$
window. Convolution-style GNNs generalize the regularity of such a
filter to irregularly connected node
pairs~\cite{GeometricDL,hamilton2017representation,Monti_2017_CVPR}.
Specifically,
the regular window is replaced by the 1-hop neighborhood of a node.

One such representative GNN is the \emph{graph convolutional network}
(GCN)~\cite{kipf2016semisupervised}. GCN maps a feature matrix $X$ to
the label matrix $Y$ (see notation introduced in
Section~\ref{sec:notation.ml}) by including the graph adjacency matrix
$A$ in the mapping:
\begin{equation}\label{eqn:GCN}
\widetilde{Y} = \softmax(\widehat{A}\cdot\relu(\widehat{A}XW^0)\cdot W^1),
\end{equation}
where $\widehat{A}$ is a normalization of $A$ through
$\widehat{A}=\widetilde{D}^{-1/2}\widetilde{A}\widetilde{D}^{-1/2}$,
$\widetilde{A}=A+I$, $\widetilde{D}=\diag(\widetilde{d}_i)$, and
$\widetilde{d}_i=\sum_j\widetilde{a}_{ij}$. Therefore, matrix products
such as $\widehat{A}X$ denote the convolution by using a 1-hop
neighborhood filter.

The following concepts are for neural networks. The functions $\relu$
and $\softmax$ are nonlinear \emph{activation functions}:
$\relu(x)=\max\{x,0\}$ is an elementwise function, while
$\softmax(x)=[e^{x_1}/c, \ldots, e^{x_d}/c]$ with $c=\sum_je^{x_j}$
is a vector function; it acts on each row if the input is a
matrix. The matrices $W^0$ and $W^1$ are called \emph{weight
matrices}. Their sizes are uniquely determined by the sizes of the
other matrices in~\eqref{eqn:GCN}. Their contents are not manually
specified but \emph{learned} through minimizing the discrepancy
between $\widetilde{Y}$ and the ground truth label matrix $Y$.

Besides GCN, the literature has seen a large number of generalizations
of lattice convolution to convolutions in the graph context, including
for example,
spectral~\cite{bruna2013spectral,henaff2015deep,kipf2016semisupervised,defferrard2016Conv}
and
spatial~\cite{PATCHY-SAN,hamilton2017inductive,velikovi2017graph,SortPool_paper18,xu2018powerful,Morris2019WL}
schemes.

GNNs such as~\eqref{eqn:GCN} essentially produce a mapping $\Phi : v
\in V \longrightarrow \Phi(v) \ \in \ \RR^c$ for every node $v$ in the
graph, if we read only one row of $\widetilde{Y}$
in~\eqref{eqn:GCN}. This mapping is almost identical to the
form~\eqref{eq:Nembed} discussed in the context of node embedding; the only
nominal difference is that the output is in a $c$-dimensional space
while the node embedding is in $d$-dimensional space. This 
difference is caused by the need for  the output $\widetilde{Y}$ to be
matched with the ground truth $Y$ that has $c$ columns ($c$
label categories); while for node embedding, the embedding dimension
$d$ may be arbitrary. However, if our purpose is not to match
$\widetilde{Y}$ with $Y$, we can adjust the number of columns of $W^1$
so that the right-hand side of~\eqref{eqn:GCN} can be repurposed for
producing a node embedding. More significantly, we may take the
elementwise minimum, the average, or the weighted average of the node
embedding vectors to form an embedding vector for the entire graph
$G=(V,E)$~\cite{duvenaud2015convolutional}. In other words, a GNN,
with a slight modification, can produce a mapping \eq{eq:Gembed} \Psi :
G \longrightarrow \Psi(G) \ \in \ \RR^{d}.  \en

Now that we see how GNNs can be utilized   to produce a graph embedding through
the mapping $\Psi$, a straightforward application of coarsening is to
use $\Psi(G_c)$ in place of $\Psi(G)$ for the classification of
$G$. This  simple idea can be made more sophisticated in two
ways. First, suppose we perform a multilevel coarsening resulting
in a sequence of increasingly coarser graphs
$G=G_0,G_1,G_2,\ldots,G_\ell$. We may concatenate the vectors
$\Psi(G_0),\Psi(G_1),\ldots,\Psi(G_\ell)$ and treat the resulting
vector as the embedding of $G$. We use the concatenation result (say
$\psi_G$) to classify $G$, though building a linear classifier that
takes $\psi_G$ as input and outputs the class label.

Second, we introduce the concept of \emph{pooling}. Pooling in neural
networks amounts to taking a min/max or (weighted) average of a group
of elements and reducing it to a single element. In the context of
graphs, pooling is particularly relevant to coarsening, since if we
recall the Galerkin projection $f^{(\ell+1)}=P_{\ell}^Tf^{(\ell)}$ in
AMG (see~\eqref{eq:AMGsys}), the interpolation matrix $P_{\ell}$ plays
the role of pooling: each column of $P_{\ell}$ defines the weights in
the averaging of nodes in the last graph into a node in the coarse
graph. Hence, in the context of GNNs, we call $P_{\ell}$ a
\emph{pooling matrix}. This matrix can be obtained directly through
the definition of a coarsening
method~\cite{defferrard2016Conv,simonovsky2017dynamic}, or it can be
unspecified but learned through
training~\cite{DiffPool-paper18,Gao2019GraphU,Lee2019SelfAttentionGP}. By
using successive poolings that form a hierarchy, recent
work~\cite{DiffPool-paper18,Lee2019SelfAttentionGP,Gao2019GraphU,knyazev2019understanding}
has shown that hierarchical pooling improves graph classification
performance.

\section{Spectral coarsening}\label{sec:spectral}
While the terms `spectral graph partitioning' and `spectral
clustering' are quite well-known, the term `spectral graph coarsening'
is less explored in the literature. There are two aspects, and
therefore possible directions, to spectral coarsening. First, it may
be desirable in various tasks to preserve the spectral properties of
the original graph, as is the case in the local variation method proposed
by~\cite{loukas2019graph}. The second aspect is that one may wish to
apply spectral information for coarsening. The method proposed
by~\cite{MSkron} falls in this category. It uses the eigenvector of
the graph Laplacian to select a set of nodes for Kron
reduction~\cite{Kron-paper13} discussed earlier.

In this section, we present an approach for the first aspect; whereas
in Section~\ref{sec:LESC.s}, we develop an approach related to the second
aspect. Regarding the first aspect, eigenvectors of the graph Laplacian
encapsulate much information on the structure of the graph. For
example, the first few eigenvectors are often used for partitioning
the graph into more or less equal partitions.  Therefore, the first
question we will ask is whether or not it is possible to coarsen a
graph in such a way that eigenvectors are `preserved.' Of course, the
coarse graph and the original one have different sizes so we will have
to clarify what is meant by this.

%------------------------------------------------------------------------------
\subsection{Coarsening and lifting}
Recall from AMG that coarsening is represented by the matrix
$P\in\mathbb{R}^{n\times n_c}$. It helps to view a coarse node as a
linear combination of a set of nodes in the original graph. Let the
$k$-th coarse node be denoted by $v_k^{(c)}$ and the set be $S_k$. The
weights are $p_{ik}$ for each $v_i\in S_k$:
\begin{equation}\label{eq:P}
    v_k^{(c)} = \sum_{v_i\in S_k} p_{ik}v_i.
\end{equation}
In the simplest case, we can let $p_{ik} = 1/|S_k|$ for all $v_i\in
S_k$. The coarse adjacency matrix is then defined as:
\begin{equation}\label{eq:CoarseMatrix}
    A_c = P^TAP.
\end{equation}
A similar framework of writing a coarsened matrix in the form of
\eqref{eq:CoarseMatrix} is adopted in \cite{loukas2019graph}. Note
that in general, $A_c$ is not binary. Here, we assign the diagonal
entries of $A_c$ to $0$ and all non-zero entries to $1$.

The original graph $G$ and its coarse graph $G_c$ have a different
number of nodes. If we wish to compare the properties of these two
graphs, it is necessary to `lift' the graph Laplacian of $G_c$ into
a matrix that has the same size as that of $G$. Let $L$ and $L_c$
denote the Laplacian of the original graph and the coarse graph,
respectively. One way to construct a matrix
$L_c$ that is guaranteed to be a graph
Laplacian is as follows~\cite{loukas2019graph,jin2020graph}:
\begin{equation}\label{eq:Lc1}
    L_c = C^T L C,
\end{equation}
where $C\in \mathbb{R}^{n\times n_c}$ is a sparse matrix with
\begin{equation}
    C_{ij}=
    \begin{cases}
        1,  & \text{node $i$ in } S_j,\\
        0,  & \text{otherwise}.
    \end{cases}
\end{equation}
Note that the matrix $P$ defined in~\eqref{eq:P} is the pseudoinverse
of $C^T$, with $C^TP = I_{n_c}$. Therefore, the lifted counterpart of
$L_c$, denoted by $L_l$, is defined as
\begin{equation}\label{eqn:lifted}
L_l = P L_c P^T,
\end{equation}
because $PC^T$ is a projector.

%------------------------------------------------------------------------------
\subsection{The projection method viewpoint}
If we extend the matrix $P$ as an orthonormal matrix, then $PP^T$ is a
projector and the lifted Laplacian defined in~\eqref{eqn:lifted}
becomes $L_l=PP^TLPP^T$. It is useful to view spectral coarsening from
the \emph{projection method}~\cite{Saad-book3} angle.

Consider an orthogonal projector $\pi$ and a general (symmetric)
matrix $A$. In an orthogonal projection method on a subspace
$\mathcal{V}$, we seek an approximate eigenpair $\tilde \lambda,
\tilde u$, where $\tilde u \ \in \ \mathcal{V}$ such that
\[ \pi (A - \tilde  \lambda I) \tilde u = 0. \]
If $V=[v_1, v_2, \cdots, v_k]$ is an orthonormal basis of
$\mathcal{V}$ and the approximate eigenvector is written as $\tilde u
= V y$, then the previous equation immediately leads to the problem
\begin{equation} 
  V^T A V y = \lambda y . \label{eq:Ritz}
\end{equation} 
The eigenvalue $\tilde \lambda $ is known as
a Ritz value  and $\tilde u $ is the associated Ritz vector.

Recall the orthogonal projector $\pi$ onto the columns of $P$; that
is, $\pi=PP^T$.  If we look at the specific case under consideration,
we notice that this is precisely what is being done and that the basis vectors
$V$ are just the columns of $P$.

When analyzing errors for projection methods, the orthogonal projector
$\pi $ represented by the matrix $P P^T$ plays a particularly
important role. Specifically, a number of results are known that can
be expressed based on the distance $\| (I - \pi) u\|$ where $u$ is an
eigenvector of $A$; see, e.g., \cite{Saad-book3}. The norm $\| (I -
\pi) u\|$ represents the distance of $u$ to the range of $\pi$ in
$\RR^n$.

%------------------------------------------------------------------------------
\subsection{Eigenvector preserving coarsening}
Based on the interpretation of projection methods, it is desirable to
construct a projector that preserves eigenvectors. We say that a given
eigenvector $u$ is exactly preserved or just `preserved' by the
coarsening if $(I-\pi) u = 0$. If this is the case then when we solve
the projected problem \eqref{eq:Ritz}, we will find that $y = P^T u$
is an eigenvector of $V^T A V$ associated with the eigenvalue
$\lambda$:
\[ V^T A V (V^T u) = V^T A \pi u = V^T A  u = \lambda V^T  u . \] 
The Ritz vector is $\tilde u = P y = P P^T u = \pi u = u $, which is
clearly an eigenvector.

What might be more interesting is the more practical situation in
which $(I-\pi) u $ is not zero but just small.  In this case, there
are established bounds \cite{Saad-book3} for the angle between the
exact and approximate eigenvectors based on the quantity $\epsilon =
\| (I-\pi) u\|$.

In what follows, we instantiate the general matrix $A$ by the graph
Laplacian matrix $L$ and consider the preservation of its
eigenvectors. From many machine learning methods (e.g., the Laplacian
eigenmap~\cite{Belkin2003}), the bottom eigenvectors of $L$ carry the
crucial information of a dataset. Thus, they are the ones that we want
to preserve.

%------------------------------------------------------------------------------
\subsection{Preserving one eigenvector}
If we want to coarsen the graph into $k$ nodes, we partition an
eigenvector $u$ into $k$ parts. Up to permutations of the elements of
$u$, let us write, in matlab notation
\[
u = \begin{bmatrix} u_1 \ ; \ u_2 \ ; \ \cdots \ ; \ u_k \end{bmatrix}.
\]
Then, we let
\[
P = \begin{bmatrix}
u_1 / \|u_1\| \\
& u_2 / \|u_2\| \\
& & \ddots \\
& & & u_k / \|u_k\|
\end{bmatrix}.
\]
Clearly, $P$ is orthonormal and satisfies $u=PP^Tu$. In other words,
the matrix $P$ so defined preserves the eigenvector $u$ of the graph
Laplacian in coarsening.

The square of an element of $u$ is called the \emph{leverage score} of
the corresponding node (see Section~\ref{sec:levr}). Then, each
$\|u_i\|^2$ is the leverage score of the $i$-th coarse node. In other
words, if a collection of nodes of the original graph is grouped into
a coarse node, then the sum of their leverage scores is the leverage
score of the coarse node.

%------------------------------------------------------------------------------
\subsection{Preserving $m$ eigenvectors}
The one-eigenvector case can be easily extended to $m$
eigenvectors. Let $U$ be the matrix of these eigenvectors; that is,
$U$ has $m$ columns, each of which is an eigenvector. We
partition $U$ similarly to  the preceding subsection, as
\[
U = \begin{bmatrix} U_1 \ ; \ U_2 \ ; \ \cdots \ ; \ U_k \end{bmatrix}.
\]
Then, we define the matrix $P$ in the following way:
\[
P = \begin{bmatrix}
P_1 \\
& P_2 \\
& & \ddots \\
& & & P_k
\end{bmatrix}
\equiv \begin{bmatrix}
U_1 R_1^{-1} \\
& U_2 R_2^{-1} \\
& & \ddots \\
& & & U_k R_k^{-1}
\end{bmatrix},
\]
where for each partition $i$, $U_i=P_iR_i$. The equality $U_i=P_iR_i$
can be any factorization that results in orthonormal matrices $P_i$
(so that $P$ is orthonormal). A straightforward choice is the QR
factorization. Alternatively, one may use the polar factorization,
where $P_i$ and $R_i$ are the unitary polar factor and the symmetric
positive definite polar factor, respectively. This factorization is
conceptually closer to the one-eigenvector case.

In contrast with the one-eigenvector case, now a collection of nodes of
the original graph is  grouped into $m$ coarse nodes, which are all
pairwise connected in the coarse graph. The total number of
nodes in the coarse graph is $m k$. Because the Frobenius norm of
$U_i$ is equal to that of $R_i$, we see that the sum of the leverage
scores of the original nodes in a partition is the same as that of the
$m$ resulting coarse nodes.

It is not hard to see that each eigenvector in $U$ is
preserved. Indeed, if $u$ is the $j$-th column of $U$, then $u = U
e_j$ and, using matlab notation,
\begin{align*}
  u &= [U_1 e_j \ ; \  U_2 e_j \ ; \  \cdots \ ; \  U_k e_j ]\\
    &= [P_1 R_1 e_j \ ; \  P_2 R_2 e_j \ ; \  \cdots \ ; \ P_k R_k e_j ]\\ 
    &\equiv [P_1 \rho_1 \ ; \  P_2 \rho_2  \ ; \  \cdots \ ; \ P_k \rho_k ]\\
  & = P [ \rho_1 \ ; \   \rho_2  \ ; \  \cdots \ ; \ \rho_k ],
\end{align*}
where we have set $\rho_i = R_i e_j$ for $i=1,\cdots, k$.  Therefore,
$u$ is in the range of $P$ and as such it will be left invariant by
the projector $\pi$: If $\rho = [ \rho_1 \ ; \ \rho_2 \ ; \ \cdots \ ;
  \ \rho_k ]$ then $\pi u = P P^T (P \rho) = P \rho = u$.

Clearly, it is not necessary to use a regular fixed and equal-sized
splitting for the rows of $U$ (and $u$); i.e., we can select any
grouping of the rows that can be convenient for, say, preserving
locality, or reflecting some clustering.

\section{Coarsening based on leverage scores}\label{sec:LESC.s}
Spectral coarsening may have desirable qualities when considering
spectral properties, but these methods face a number of practical
difficulties.  Among them is the fact that the coarsened graph tends
to be dense.  For this reason, spectral methods will be invoked mostly
as a tool to provide an ordering of the importance of the
nodes---which will in turn be used for defining a traversal order in
other coarsening approaches (e.g., HEM). This has been a common theme
in the literature \cite{HARP-paper18,jin2020graph,loukas2019graph}.

%------------------------------------------------------------------------------
\subsection{Leverage scores}\label{sec:levr}
Let $A$ be a general matrix and let $U$ be an orthonormal matrix,
whose range is the same as that of $A$. The leverage
score~\cite{Leverage} of the $i$-th row of $A$ is defined as the
squared norm of the $i$-th row of $U$:
\begin{equation} \label{eq:gammai}
  \eta_i = \|U_{i,:} \|_2^2 .
\end{equation}
Clearly, the leverage score is invariant to the choice of the
orthonormal basis of the range of $A$.

Leverage scores defined in the form~\eqref{eq:gammai} have been used
primarily for (rectangular) matrices $A$ that represent data. In these
cases, the columns of $U$ are the dominant singular vectors of $A$ and
the matrix $\pi=UU^T$ is an orthogonal projector that projects a vector
in $\RR^n$ onto the span of $U$. The vector of leverage scores,
$\eta$, is the diagonal of this projector. This quantity appears also
in a different context in quantum physics, where the index $i$
represents a location in space, $\eta_i$ represents the electron
density in position $i$, and the projector $\pi$ is the density matrix
in the idealistic case of zero temperature. See Section 3.4
of~\cite{RMartin-book}.

%------------------------------------------------------------------------------
\subsection{Leverage-score coarsening (LESC)}\label{sec:LESC}
For coarsening methods such as Algorithm~\ref{alg:HEM}, the traversal
order in the coarsening process can have a major impact on the quality
of the results. Instead of the heavy edge matching strategy, we now
consider using leverage scores to measure the importance of a
node. Exploiting what we know from spectral graph theory, we will use
the bottom eigenvectors of the graph Laplacian $L$ to form
$U$. However, we find that in many cases, the traversal order is
sensitive to the number of eigenvectors, $r$.  To lessen the impact of
$r$ on the outcome, we weigh individual entries $U_{ik}$
in~\eqref{eq:gammai} by using the eigenvalues $\lambda_1, \ldots,
\lambda_r$ of $L$. Specifically, we define
\begin{equation} \label{eq:etair}
  \eta_i = \sum_{k=1}^{r}(e^{-\tau\lambda_k}U_{ik})^2,
\end{equation}
where $\tau$ represents a decay factor of the weights. This leads to
a modification of HEM that is based on
 leverage scores~\eqref{eq:etair} which we call 
\emph{leverage score coarsening}, or LESC for short.

Algorithm~\ref{alg:LESC} describes the LESC procedure. Its main
differences from HEM (Algorithm~\ref{alg:HEM}) lie in (i) the
traversal order and (ii) the handling of singletons. While HEM
proceeds by the heaviest edge, LESC scans nodes in decreasing $\eta$
values. At the beginning of each for-loop, LESC selects the next
unvisited node $i$ with the highest leverage score and selects from
its unvisited neighbors a node $j$, where the edge $(i,j)$ has the
heaviest edge weight, to create a coarse node.

The way in which LESC handles the singletons is elaborated in lines
14--25 of Algorithm~\ref{alg:LESC}. During the traversal, we append a
singleton to a set named $Single$. Depending on the desired coarse
graph size $n_c$, there are two ways to assign parents to the
singletons: lines 20--22 handle a real singleton and lines 23--25
handle a leftover singleton. This extra step outside HEM better
preserves the local structure surrounding high-degree nodes, as well
as the global structure of the graph.

\begin{algorithm}[h]
  \caption{Leverage-Score Coarsening (LESC)}
  \begin{algorithmic}[1]
    \State {\bfseries Input:} Weighted graph $G=(V,E,A)$; leverage score $\eta$; coarse graph size $n_c$.
    \State {\bfseries Output:} Coarse nodes; $Prnt$ list
    \State {\bfseries Init:} $Prnt(i)=0$ $\forall i\in V$, $new=0$, $Single = \emptyset$
    \For{$\max$-score {\bfseries to} $\min$-score node $i$}
    \If{$i$ has no neighbor}
    \State $new = new + 1$; $Prnt(i)=new$
    \Else
    \For{max {\bfseries to} min edge $(i,j)$}
    \If{$Prnt(j)==0$}
    \State $new=new+1$
    \State $Prnt(i)=Prnt(j)=new$
    \EndIf
    \EndFor
    \If{$Prnt(i)==0$}
    \State Add $i$ to $Single$ set
    \EndIf
    \EndIf
    \EndFor
    \State Randomly shuffle $Single$ set
    \For{the first $n_c-new$ nodes $v$ {\bfseries in} $Single$}
    \State $new = new + 1$; $Prnt(v)=new$
    \EndFor
    \For{remaining nodes $v$ {\bfseries in} $Single$}
    \State $Prnt(v)=Prnt(j)$ where $j=\argmax_i(A_{i,v})$
    \EndFor
  \end{algorithmic}
  \label{alg:LESC}
\end{algorithm}

\begin{figure}
\captionsetup[subfigure]{justification=centering}
  \centering
  \begin{subfigure}[b]{0.3\textwidth}
    \centering
    \includegraphics[width=\textwidth]{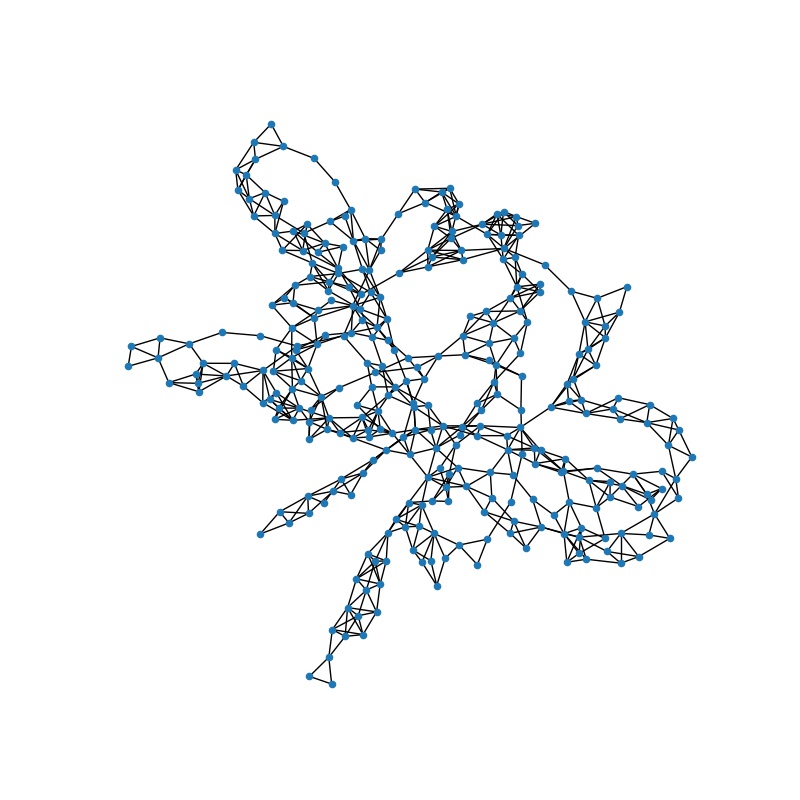}
    \caption{Original\\ 312 nodes, 761 edges}
  \end{subfigure}
  \begin{subfigure}[b]{0.3\textwidth}
    \centering
    \includegraphics[width=\textwidth]{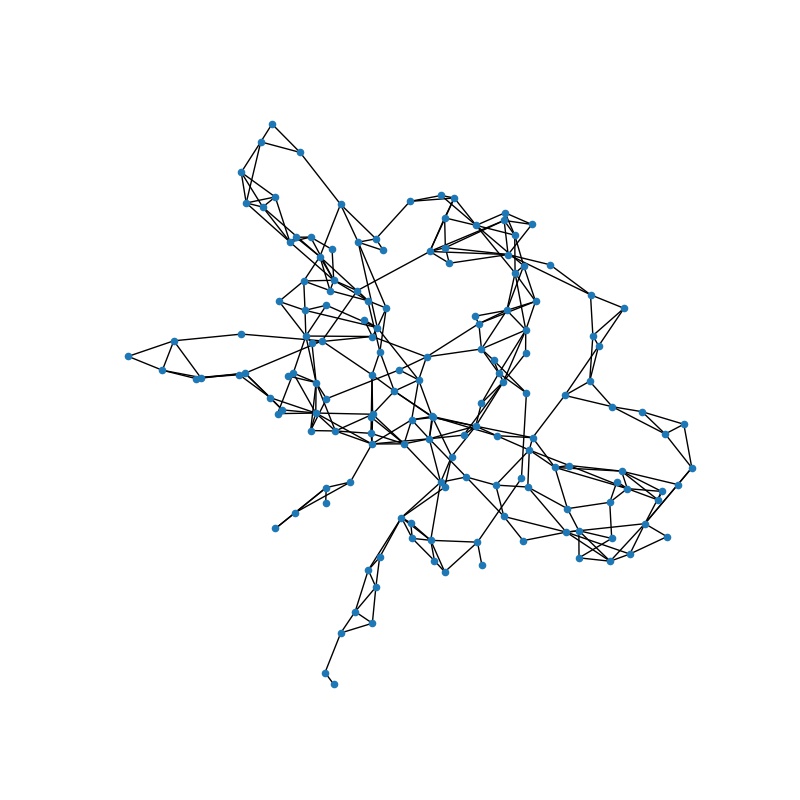}
    \caption{HEM \\ 156 nodes, 340 edges}
  \end{subfigure}
  \begin{subfigure}[b]{0.3\textwidth}
    \centering
    \includegraphics[width=\textwidth]{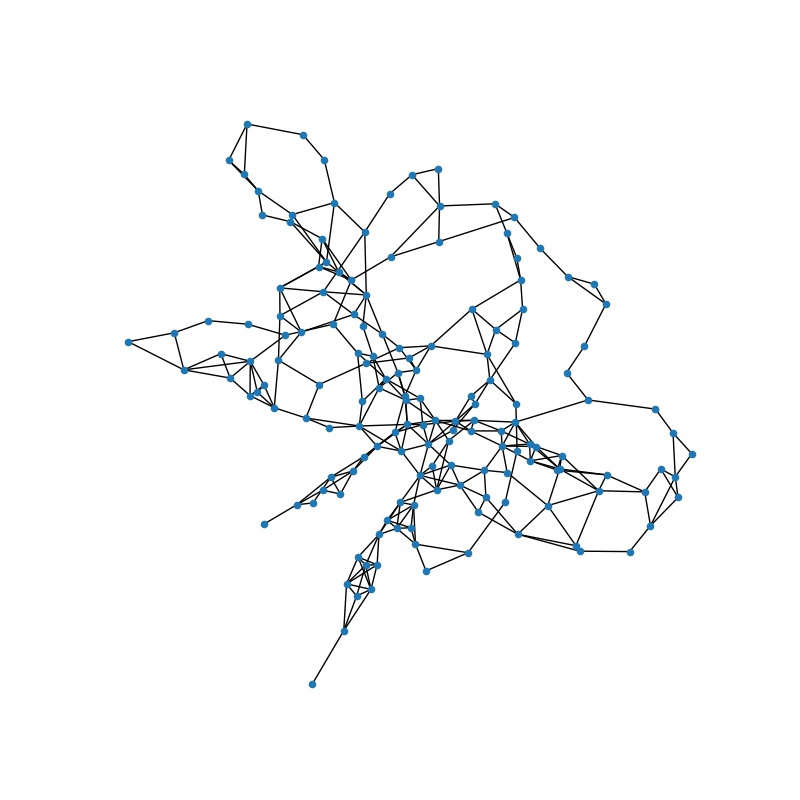}
    \caption{LV \\ 156 nodes, 321 edges}
  \end{subfigure}
    \begin{subfigure}[b]{0.3\textwidth}
    \centering
    \includegraphics[width=\textwidth]{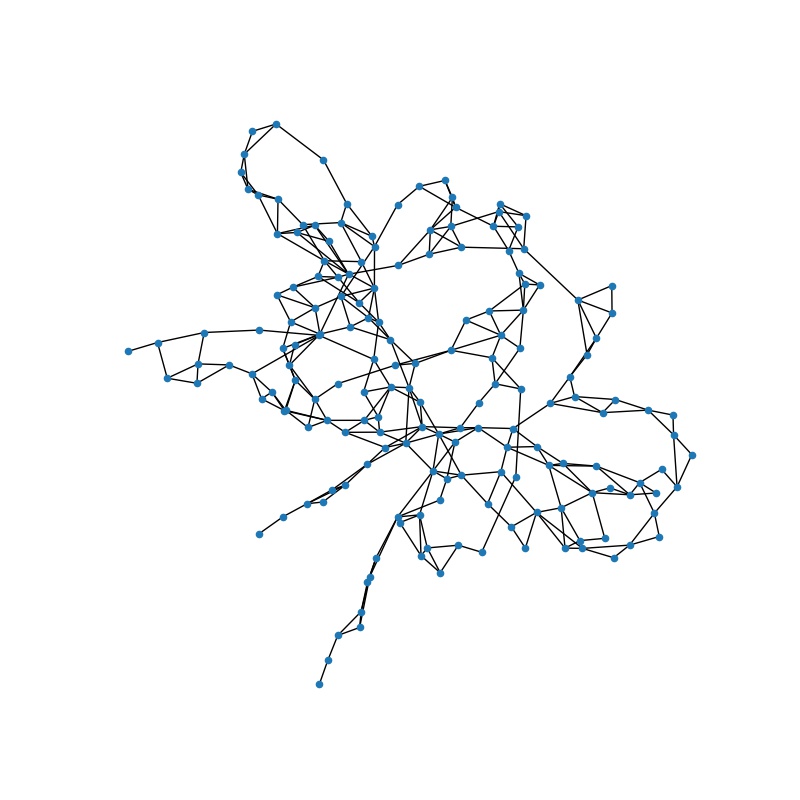}
    \caption{algebraic distance\\ 171 nodes, 327 edges}
  \end{subfigure}
  \begin{subfigure}[b]{0.3\textwidth}
    \centering
    \includegraphics[width=\textwidth]{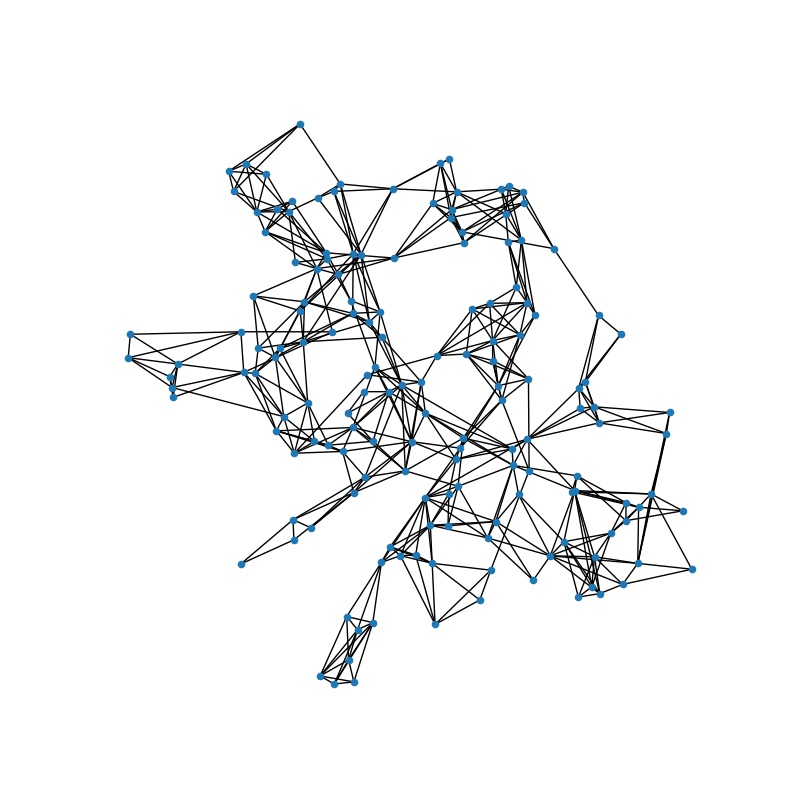}
    \caption{Kron \\ 156 nodes, 485 edges}
  \end{subfigure}
  \begin{subfigure}[b]{0.3\textwidth}
    \centering
    \includegraphics[width=\textwidth]{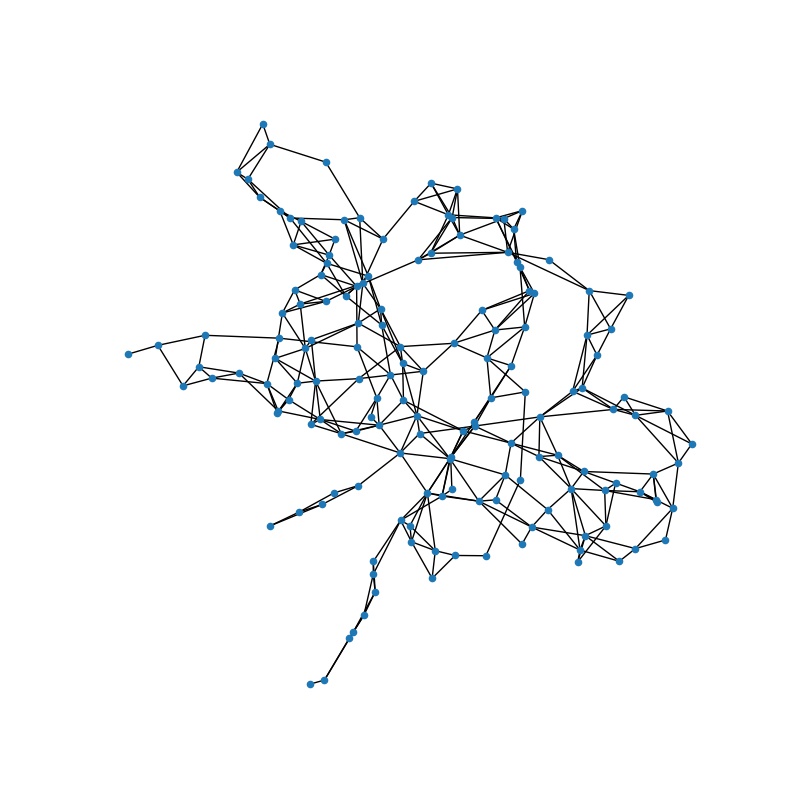}
    \caption{LESC \\ 156 nodes, 362 edges}
  \end{subfigure}
  \caption{Visualization of a graph and the coarsened graphs obtained
    by different methods.}
  \label{fig:DDvisual}
\end{figure}

For an illustration, we visualize the coarse graphs produced by the
following five coarsening methods: HEM, local variation
(LV)~\cite{loukas2019graph}, algebraic distance, Kron reduction, and
LESC. The original graph is selected from the D\&D protein data set;
see Section~\ref{sec:exp} for a detailed description. The coarse
graphs are shown in Figure \ref{fig:DDvisual}. We observe from the
figure that the global structure of the graph is well preserved in
each coarse graph. Moreover, the connection chains of the nodes are
well preserved. Note also that, as expected, the coarse graph from
Kron reduction is denser than the other coarse graphs.

%%%%====

%------------------------------------------------------------------------------
\subsection{Interpretations}\label{sec:interp}
Leverage scores defined in~\eqref{eq:gammai} have been used primarily
for matrices that represent data. The form used by us,
\eqref{eq:etair}, stabilizes the ordering of the scores when the
number $r$ of dominant eigenvectors varies. When $\tau$ or $r$ is
large, there is little difference between the value~\eqref{eq:etair}
and the following one that uses all eigenvectors:
\begin{equation} \label{eq:etai}
  \eta_{i} = \sum_{k=1}^{n}(e^{-\tau\lambda_k}U_{ik})^2.
\end{equation}
The form~\eqref{eq:etai} can lead to interesting interpretations and
results.

First, observe that if we denote by $e_i$ the $i$-th column of the
identity matrix, then
\begin{equation} \label{eq:etai1}
  \eta_{i} = \sum_{k=1}^{n} e^{-2\tau\lambda_k} |U_{ik}|^2 = e_i^T e^{-2\tau L} e_i.
\end{equation}
That is, $\eta_i$ is nothing but the $i$-th diagonal entry of the
matrix $H \equiv \exp (- 2\tau L)$. If the adjacency matrix $A$ is
doubly stochastic, then $L = I - A$ and $H = \exp (- 2\tau L) = \exp
(- 2\tau I + 2\tau A) = e^{- 2\tau} \times \exp(2\tau A) $. Therefore,
since $A$ has nonnegative entries, so does $H$. Then, $H$ is a
stochastic matrix (in fact, doubly stochastic because of symmetry). To
see this, we have $L\ones=0$ and thus by the Taylor series of the
matrix exponential, $H\ones=\ones$. Now that $H$ is a stochastic
matrix, the leverage score $\eta_i$ ($i$-th diagonal entry of $H$)
represents the self-probability of state $i$.

There exists another interpretation from the \emph{transient solutions
of Markov chains}; see Chapter 8 of~\cite{Billy-book}. In the
normalized case, the negative Laplacian $-L$ plays the role of the
matrix $Q$ in the notation of continuous time Markov chains (see
Section 1.4 of~\cite{Billy-book}). Given an initial probability
distribution $\pi(0) \ \in \ \RR^{1 \times n}$, the transient solution
of the chain at time $t$ is $\pi (t) = \pi(0) \exp ( t Q) = \pi(0)
\exp ( -t L) $. If $\pi(0) = e_i^T $, then $\pi (t)$ carries the
probabilities for each state at time $t$. In particular, the $i$-th
entry (which coincides with the leverage score $\eta_i$ if $t=2\tau$)
is the probability of remaining in state $i$.

%------------------------------------------------------------------------------
\subsection{Alternative definition}
\begin{figure}[h]
  \centering
    \includegraphics[width=\columnwidth]{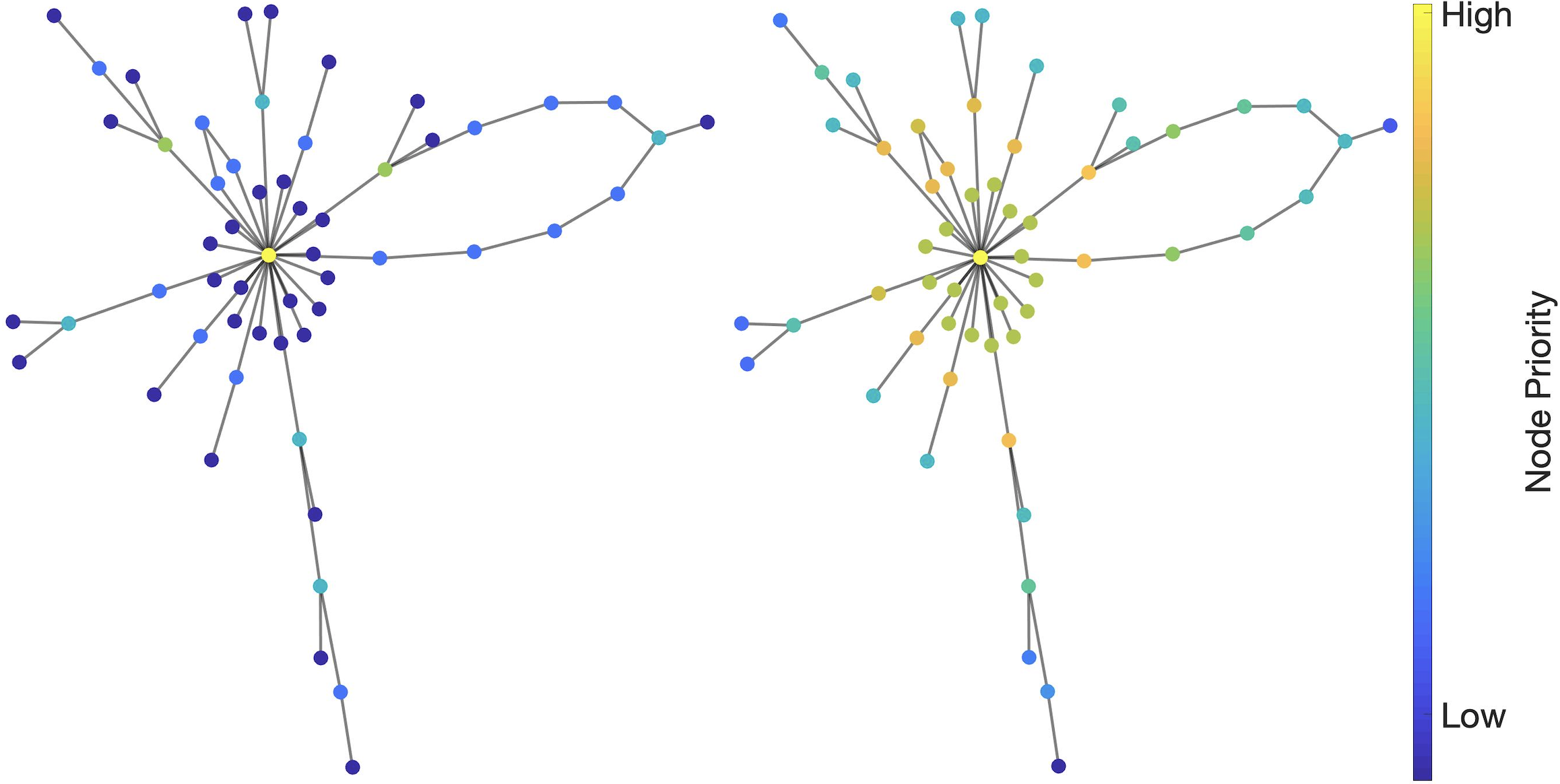}
  \caption{Traversal order of HEM (left) and LESC (right) on an
    unweighted graph.}
  \label{fig:oder}
\end{figure}

The definition of $\eta_i$ in~\eqref{eq:etair} modifies the standard
leverage score by using decaying weights, to reduce sensitivity of the
number of eigenvectors used. In principle, any decreasing function of
eigenvalues can be used to get distinguishable leverage scores of a
Laplacian.  We consider the following alternative, which is related to
the pseudoinverse of the Laplacian:
\begin{equation} \label{eq:etair1}
  \eta_i = \sum_{j=2}^{n}\left(\frac{1}{\sqrt{\lambda_j}}U_{ij}\right)^2.
\end{equation}
Several points are worth noting. First, weighted leverage scores
emphasize eigenvectors corresponding to small non-zero
eigenvalues. Hence, weighted leverage scores reveal the contribution
of nodes to the global structure. Second, a smaller weighted leverage
score indicates a higher topological importance of a node. Third,
calculating the complete set of eigenvectors of $L$ is
expensive. Given a parameter $r$, we can further define $r$-truncated
weighted leverage scores using only $r$ eigenpairs:
\begin{equation}\label{eq:eta}
  \eta_i = \sum_{j=2}^r\left(\frac{1}{\sqrt{\lambda_j}}U_{ij}\right)^2.
\end{equation}
For simplicity, we refer to these numbers as leverage scores of $L$,
and use $\eta = [\eta_1,\cdots,\eta_n]$ to denote them. A visual example of
using $\eta$ to define the traversal order in Algorithm~\ref{alg:LESC}
is given in Figure~\ref{fig:oder}.

The definition~\eqref{eq:etair1} has a direct connection with the
pseudoinverse of the Laplacian. In particular, the vector $\eta$ is
equal to the diagonal of $L^\dagger$. To see this, we first notice
that $L$ and $L^\dagger$ have the same set of eigenvectors, and
nontrivial eigenvalues are reciprocals of each other.
We then write  $L^\dagger $ as
$L^\dagger= U\Sigma^\dagger U^T$, where $\Sigma^\dagger$ is
a diagonal matrix with $0<1/\lambda_n\le\cdots\le1/\lambda_2$ on the
diagonal. Diagonal entries of $\Sigma^\dagger $ are  
 non-negative, so we can write
 \eq{eq:Ldag}
 L^\dagger =
 U\sqrt{\Sigma^\dagger}\left(U\sqrt{\Sigma^\dagger}\right)^T,
 \en
 from which we get
\[ 
    L_{ii}^\dagger = \sum_{j=2}^{n}\frac{U_{ij}}{\sqrt{\lambda_j}}\frac{U_{ij}}{\sqrt{\lambda_j}}=\eta_i.
\]
The pseudoinverse of $L$ has long been used to denote node
importance. The article \cite{van2017pseudoinverse} provides a rather
detailed description of the link between $L^\dagger$ and the various
definitions of network properties. The effective resistance
distance between  nodes $a$ and $b$ is given by
$\omega_{ab} = L^\dagger _{aa} + L^\dagger_{bb} - 2 L^\dagger_{ab}$.
%% A nonzero entry of $L^\dagger$ defines the well-known
%% effective resistance distance~\cite{teng2010laplacian}.
The trace of
$L^\dagger$ defines a graph metric called \emph{effective graph
resistance}~\cite{ellens2011effective}, which is related to random
walks~\cite{chandra1996electrical} and the betweenness
centrality~\cite{newman2005measure}.
In addition, the columns $s_j$ of the matrix $U \sqrt{\Sigma^\dagger}$ in
\nref{eq:Ldag} have a particular significance.
The squared distance between two columns
$s_a$ and $s_b$ is equal to $\omega_{ab}$  and based on this
notion of distance, the authors of \cite{ranjan2013geometry} propose to use
$(L_{ii}^\dagger)^{-1}$ as a measure of the topological centrality of
a node $i$. The smaller the distance, the higher its topological
centrality. This metric has been used to quantify the roles of nodes
in independent networks~\cite{shin2014cascading}. Therefore, by using
$\eta$, LESC prioritizes nodes with high importance with respect to this metric.

The leverage score vector $\eta$ is also related to the change of the
Laplacian pseudoinverse when merging a pair of nodes in coarsening. We
follow the work by~\cite{hermsdorff2019unifying} to elucidate this. To
simplify the comparison between two matrices with different sizes,
consider the following perspective: during coarsening, instead of
merging a pair of nodes and reducing the graph size by one, we assign
the corresponding edge with an edge weight $+\infty$. To avoid
possible confusion with $L_c$, we use $L_{\infty}$ to denote the
coarse graph Laplacian (the Laplacian of the $+\infty$-weighted
graph). Suppose we assign an edge $e(v_i,v_j)$ with the $+\infty$
weight, then the difference between $L$ and $L_{\infty}$ is
\begin{equation}
\Delta L = L_{\infty} - L = w b_eb_e^T, \quad(w = +\infty)
\end{equation}
where $b_e$ is defined in~\eqref{eq:b}. Then, the change in $L^{\dagger}$ is given by the Woodbury matrix identity~\cite{goldfarb1972modification}:
\begin{equation}
    \Delta L^{\dagger} = -\frac{w}{1+ w b_e^TL^{\dagger}b_e}L^{\dagger}b_eb_e^T L^{\dagger}
    = -\frac{1}{b_e^TL^{\dagger}b_e}L^{\dagger}b_eb_e^T L^{\dagger}. \quad (w = +\infty)
\end{equation}
Thus, the magnitude of $\Delta L^\dagger$ can be defined by the Frobenius norm:
\begin{equation}\label{eq:MagL}
    ||\Delta L^{\dagger}||_F^2 = \frac{b_e^T L^\dagger L^\dagger b_e}{b_e^T L^\dagger b_e}.
\end{equation}
The following result bounds the magnitude of $\Delta L^\dagger$ by using
leverage scores.

\begin{proposition}\label{prop:main}
Let the graph be connected. The magnitude of the difference between
$L^\dagger$ and $L_\infty^\dagger$ caused by assigning the $+\infty$
edge weight to an edge $e(i,j)$ is bounded by
\begin{equation*}
||\Delta L^\dagger||_F^2\le \kappa(L)(L_{ii}^\dagger + L_{jj}^\dagger),
\end{equation*}
where $\kappa$ denotes the effective condition number (i.e., the
largest singular value divided by the smallest nonzero singular
value).
\end{proposition}

\begin{proof}
Let $L^{\dagger}=U\Lambda U^T$ be the spectral decomposition of
$L^{\dagger}$ with eigenvalues sorted nonincreasingly: $\mu_1 \ge
\cdots \ge \mu_{n-1} > 0 = \mu_n$. Further, let
$x=\Lambda^{1/2}U^Tb_e$.  Then,
\[
   \frac{b_e^TL^{\dagger}L^{\dagger}b_e}{b_e^TL^{\dagger}b_e}
=\frac{x^T\Lambda x}{x^Tx}
\le \mu_1.
\]
Now consider a lower bound and an upper bound of
$b_e^TL^{\dagger}b_e$. Since $b_e$ is orthogonal to the vector of all
ones (an eigenvector associated with $\mu_n$), we have
\[
\left(\frac{b_e}{\sqrt{2}}\right)^TL^{\dagger}\left(\frac{b_e}{\sqrt{2}}\right) \ge \min_{\|y\|=1}\{y^TL^{\dagger}y \mid y^T\ones=0 \} = \mu_{n-1}.
\]
On the other hand, note that
$b_e^TL^{\dagger}b_e=L^{\dagger}_{ii}+L^{\dagger}_{jj}-2L^{\dagger}_{ij}$
and that $(L^{\dagger}_{ij})^2 \le L^{\dagger}_{ii}L^{\dagger}_{jj}$
(by positive semidefiniteness). Then,
\[
b_e^TL^{\dagger}b_e
\le \left(\sqrt{L^{\dagger}_{ii}}+\sqrt{L^{\dagger}_{jj}}\right)^2
\le 2(L^{\dagger}_{ii} + L^{\dagger}_{jj}).
\]
Invoking both the lower bound and the upper bound, we obtain
\[
\frac{b_e^TL^{\dagger}L^{\dagger}b_e}{b_e^TL^{\dagger}b_e}
\le \mu_1
\le \mu_1 \frac{2(L^{\dagger}_{ii} + L^{\dagger}_{jj})}{2\mu_{n-1}}.
\]
Then, by noting that $\mu_1/\mu_{n-1}$ is the effective
condition number of $L^{\dagger}$ (as well as $L$), we conclude the
proof.
\end{proof}
\subsection{Experimental results}\label{sec:exp}
Here, we show an experiment to demonstrate the effective use of LESC
to speed up the training of GNNs. We focus on the task of \emph{graph
classification} that assigns a label to a graph (see
Section~\ref{sec:GNN}). As mentioned in that section, we classify the
coarsened graph rather than the original graph. We use four
hierarchical GNNs (SortPool~\cite{SortPool_paper18},
DiffPool~\cite{DiffPool-paper18},
TopKPool~\cite{Gao2019GraphU,Cangea2018TowardsSH}, and
SAGPool~\cite{Lee2019SelfAttentionGP}) to perform the task and
investigate the change of training time and prediction accuracy under
coarsening.

We use three data sets for evaluation:
D\&D~\cite{dobson_distinguishing_2003}, REDDIT-BINARY (REBI), and
REDDIT-MULTI-5K (RE5K)~\cite{DeepGraphKernels}. The first is a protein
data set and the label categories are protein functions. Each protein
is represented by a graph, where nodes represent amino acids and they
are connected if the two acids are less than six Angstroms apart. The
last two are are social network data sets collected from the online
discussion forum Reddit. Each discussion thread is treated as one
graph, in which a node represents a user and there is an edge between
two nodes if either of the two users respond to each other's
comment. Statistics of the data sets are given in
Table~\ref{tab:statistics}.

\begin{table}[htbp]
  \centering
  \caption{Dataset statistics.}
  \label{tab:statistics}
    \begin{tabular}{c c c c }
      \hline
                  & D\&D     & REBI & RE5K \\
      \hline
      \#graphs    & 1178   & 2000   & 4999     \\
      \#classes   & 2      & 2      & 5           \\
      Avg.\#nodes & 284.32 & 429.63 & 508.52  \\
      Avg.\#edges & 715.66 & 497.75 & 594.87 \\
      \hline
    \end{tabular}
\end{table}

\begin{figure}[htb]
  \centering
    \includegraphics[width=\textwidth]{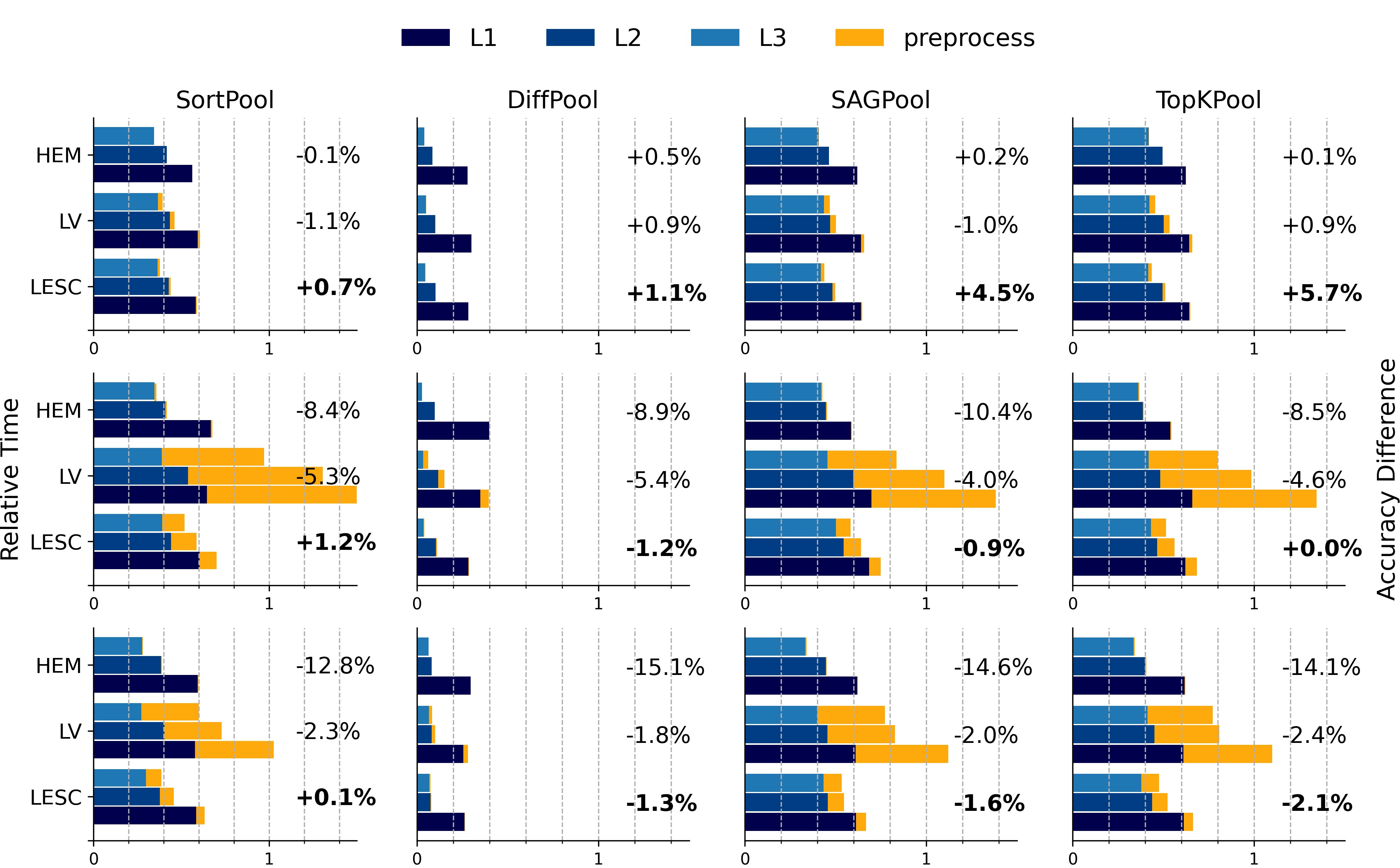}
  \caption{Relative time and accuracy difference on D\&D (top row), REBI (middle row), and RE5K (bottom row). Four classification methods (SortPool, DiffPool, SAGPool, and TopKPool) and three coarsening methods (HEM, LV, and LESC) are considered. For each classification method, we normalize the run time by that of the original method. Each run time consists of two parts: coarsening and GNN training with 10-fold cross validation. For each coarsening method, run times are reported for a number of coarsening levels. The highest accuracy achieved by each coarsening method is annotated on the right of the bar chart through differencing from that of the original graph (coarsening minus original). Positive values indicate accuracy increase. The best case for each chart is highlighted in boldface. }
  \label{fig:Time}
\end{figure}

Figure~\ref{fig:Time} summarizes the time (bars) and accuracy
(percentages) results. Each column is for one GNN and each row is for
one data set. Inside a panel, three coarsening methods are compared,
each using three coarsening levels.

When comparing times, note that applying coarsening to graph
classification incurs two costs: the time to perform coarsening (as
preprocessing) and the time for training. Therefore, we normalize the
overall time by that of training a GNN without using
coarsening. Hence, a relative time $<1$ indicates improvement. In
fact, the relative time is just the reciprocal of the speedup.

Here are a few observations regarding training times.  First, for all
three data sets, time reduction is always observed for HEM and
LESC. Second, for these two methods, the coarsening time is almost
negligible compared with the training time, whereas LV incurs a
substantial  overhead for coarsening in several cases. In LV, the
coarsening time may even dominate the training time (see REBI and
RE5K). Discounting coarsening, however, LV produces coarse graphs that
lead to reduced training times. Third, in general, the deeper the
levels, the more significant is the time reduction. The highest
speedup for LESC, which is approximately 30x, occurs for REBI in three
levels of coarsening.

To compare  prediction accuracies, we compute the relative change (in
\%) in accuracy and annotate it on the right side of each panel in
Figure~\ref{fig:Time}. We see that LESC achieves the best performance
among all three coarsening methods on all three datasets. It also
improves accuracy on a few of the GNNs while  for the others it
produces an accuracy that is quite close to that achieved by not using
coarsening. In other words, coarsening does not negatively impact
graph classification overall.

\section{Concluding remarks}
This survey (with new results) focused on graph coarsening techniques with a
goal of showing how some common ideas have been utilized in two different
disciplines while also highlighting methods that are specific to some
applications.  The recent literature on methods that employ graphs to model data
clearly indicates that the general method is likely to gain importance. This is
only natural because the graphs encountered today are becoming large and
experiments show that if employed with care, coarsening does not cause a big
degradation in the performance of the underlying method.  As researchers in
numerical linear algebra and scientific computing are increasingly turning their
attention to problems related to machine learning, graph based methods, and
graph coarsening in particular, are likely to play a more prominent role.

\bibliographystyle{siam} 
\bibliography{strings,refs,ilu,GNN2,graph,saad,mlevel,gsparsification}

\end{document}